\documentclass{article}

\usepackage[square,numbers]{natbib}

\usepackage{hyperref}       % hyperlinks
\hypersetup{
    colorlinks,
    linkcolor={black!50!black},
    citecolor={black!50!black},
    urlcolor={blue!80!black}
}
\usepackage{url}            % simple URL typesetting
\usepackage[preprint]{neurips_2024}

\usepackage[utf8]{inputenc} % allow utf-8 input
\usepackage[T1]{fontenc}    % use 8-bit T1 fonts

\usepackage{booktabs}       % professional-quality tables
\usepackage{amsfonts}       % blackboard math symbols
\usepackage{nicefrac}       % compact symbols for 1/2, etc.
\usepackage{microtype}      % microtypography
\usepackage{xcolor}         % colors
\usepackage{wrapfig,lipsum,booktabs}
\usepackage{graphicx}
\usepackage{subfig}

\usepackage{algorithm}% http://ctan.org/pkg/algorithms
\usepackage{algorithmic}

\usepackage{caption}
\usepackage{subcaption}

\usepackage{amsmath}
\usepackage{amssymb}
\usepackage{mathtools}
\usepackage{amsthm}

\usepackage[capitalize,noabbrev]{cleveref}

\usepackage{url}            % simple URL typesetting
\usepackage{nicefrac}       % compact symbols for 1/2, etc.
\usepackage{microtype}      % microtypography
\usepackage{xcolor}         % colors

\usepackage{multirow}
\usepackage{amsfonts}
\usepackage{bbm}
\usepackage{arydshln}
\usepackage{etoolbox,xspace}
\usepackage{pifont}% http://ctan.org/pkg/pifont

\usepackage{float}

\newcommand{\cD}{\mathcal{D}}

\newcommand{\cL}{\mathcal{L}}

\newcommand\norm[1]{\left\lVert#1\right\rVert}

\newcommand{\bp}{\mathbf{p}}

\newcommand{\bk}{\mathbf{k}}

\newcommand{\bx}{\mathbf{x}}

\newcommand{\by}{\mathbf{y}}
\newcommand{\bv}{\mathbf{v}}

\newcommand{\bq}{\mathbf{q}}

\newcommand{\btheta}{\boldsymbol{\theta}}

\newcommand{\ours}{LoRSU\xspace}

\newcommand{\argmax}{\mathop{\mathrm{argmax}}\limits} 

\newcommand{\xmark}{\ding{55}}%

\theoremstyle{plain}
\newtheorem{theorem}{Theorem}[section]

\newtheorem{lemma}[theorem]{Lemma}
\newtheorem{corollary}[theorem]{Corollary}
\theoremstyle{definition}
\newtheorem{definition}[theorem]{Definition}

\theoremstyle{remark}
\newtheorem{remark}[theorem]{Remark}

\newcommand{\myparagraph}[1]{\noindent\textbf{#1}}
\title{Imperfect Vision Encoders: Efficient and Robust Tuning for Vision-Language Models}

\author{%
  Aristeidis Panos \\%\thanks{Use footnote for providing further information
 University of Cambridge\\
  \And
   Rahaf Aljundi\\
  Toyota Motor Europe \\
 \AND
Daniel Olmeda Reino \\
   Toyota Motor Europe \\
   \And
Richard E Turner \\
 University of Cambridge
}

\begin{document}

\maketitle

\begin{abstract}
Vision-language models (VLMs) demonstrate impressive capabilities in visual question answering and image captioning, acting as a crucial link between visual and language models. However, existing open-source VLMs heavily rely on pretrained and frozen vision encoders (such as CLIP). Despite CLIP’s robustness across diverse domains, it still exhibits non-negligible image understanding errors. These errors propagate to the VLM responses, resulting in sub-optimal performance. In our work, we propose an efficient and robust method for updating vision encoders within VLMs. Our approach selectively and locally updates encoders, leading to substantial performance improvements on data where previous mistakes occurred, while maintaining overall robustness. Furthermore, we demonstrate the effectiveness of our method during continual few-shot updates. Theoretical grounding, generality, and computational efficiency characterize our approach.
\end{abstract}
\section{Introduction}\label{sec:introduction}

% VLMs, relying on pretrained vision encoders, demonstrate impressive performance across question-answering and image-captioning benchmarks. CLIP, a widely deployed vision transformer model, exhibits robustness to domain shifts. However, our evaluation revealed limitations in performance. Specifically, when tested on an action recognition dataset featuring various home actions, CLIP exhibited moderate performance and significant confusion. Further testing with other VLMs using CLIP as the vision encoder also demonstrated poor results. These findings underscore the need for continuous model improvements to address these imperfections.
Large Language Models (LLMs) have significantly transformed the landscape of natural language understanding and generation, revolutionizing a wide range of domains and applications. These advancements bring us one step closer to creating assistants with reliable intelligence levels. Given that vision and visual understanding play a crucial role in intelligent agents expected to operate in the real world, Vision and Language Models (VLMs) have emerged. These models either incorporate embeddings from vision-only models or are trained end-to-end with both vision and language input. Remarkably, VLMs consistently achieve impressive performance across question-answering and image-captioning benchmarks. We refer to \cite{ghosh2024exploring} for a recent survey on VLMs. Approaches that rely on pretrained vision encoders, typically use variants of the CLIP model, which is kept frozen in the vision language binding process. CLIP~\cite{radford2021learning}, a widely deployed vision and text transformer, stands out for its robustness to domain shifts and outstanding capabilities of recognizing large scale of objects, scenes and actions. However, our evaluation reveals certain limitations in CLIP’s performance. Specifically, when tested on an action recognition dataset featuring various simple actions with moderate image quality, CLIP exhibits substandard performance and seems easily confounded by the image content. Other works~\cite{liu2024visual,zhu2023minigpt,chen2023minigpt,li2023blip} reveal similar shortcomings of CLIP for particular use cases. Surprisingly, CLIP and CLIP-enabled VLMs demonstrate a highly accurate understanding of images generated by AI, representing similar actions. These findings underscore weaknesses in visual understanding of CLIP, specially on challenging and previously unseen domains, and prompts the need for continuous model improvements to address these imperfections.

In order to enable VLMs to adapt to OOD data, we envision a realistic scenario where the model can be updated efficiently with minimal computational resources while maintaining its strong performance on other data and domains. In other words, we aim to correct mistakes effectively while preserving existing knowledge.

Given the composite nature of VLMs, which combine vision encoders and language models, the crucial question arises: Which components are better suited for targeted updates? To address this, we conducted separate fine-tuning experiments on  the vision encoder and the Language Model (LLM) using a dataset where the VLM exhibited numerous mistakes. The results were intriguing: \textit{separately} updating the vision encoder significantly improved the performance on the specific data of interest achieving even better accuracy than updating the LLM. Updating the vision encoder is more efficient as it contains far fewer parameters than the language model and can improve different VLMs that build on it.
Our findings suggest that separately updating the vision encoder provides a more robust alternative to LLM updates when visual shift is the primary source of errors. 
%\rahaf{not sure if we will have that} For further insights, we refer to the detailed experiments on when to update the vision encoder.

% Despite the effectiveness of vision encoder updates, continuous and frequent updates can lead to performance deterioration. Therefore, we recognize the need for not only efficient updates but also localized to the data at hand in order to limit degradation in unrelated areas of knowledge. While parameter-efficient fine-tuning methods offer efficiency, they often sacrifice the preservation of unrelated model knowledge (updates locality) similar to full finetuning, as shown before by \citet{zhang2023overcoming} and as demonstrated in our experiments. For instance, LoRa~\cite{hu2021lora}’s low-rank updates still alter all model parameters, resulting in performance decline upon multiple updates.

% In order to perform localized updates, we speculate that only parameters relevant to the task at hand should be modified while keeping rest of model's parameters intact. A similar direction is followed in Language model editing~\cite{wang2023knowledge}, however approaches are usually specific to the nature factual knowledge updates and cannot be easily applied to vision related updates. 
\begin{figure}[t]
%\vspace{-0.5cm}
    \centering
   \includegraphics[width=0.9\textwidth]{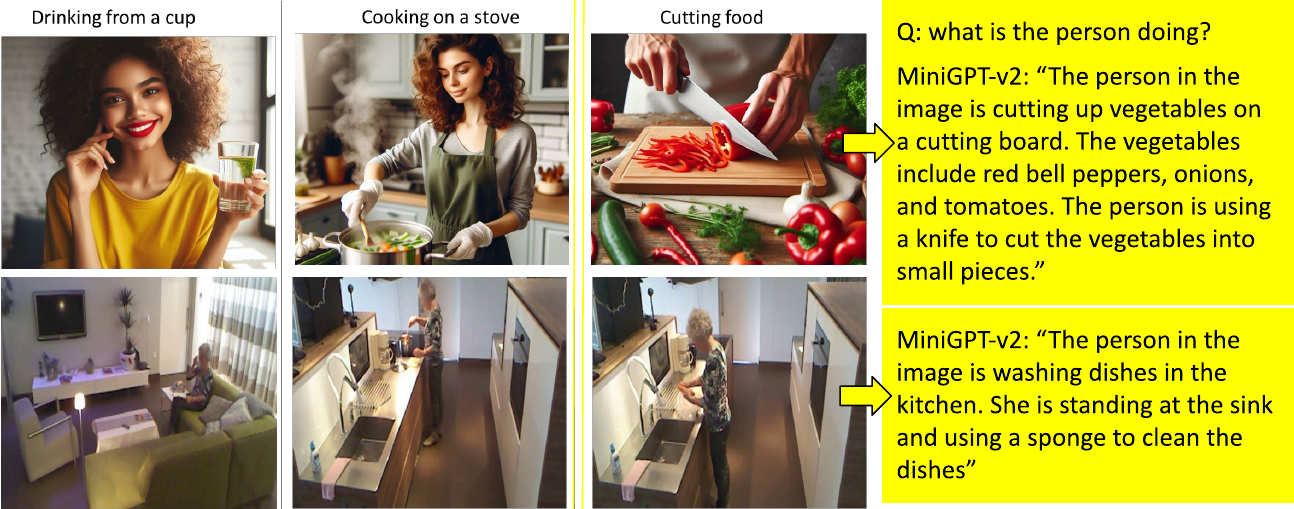}
\captionof{figure}{\label{fig-Dalle-tsi}\small Left: samples  of images from TSI dataset compared to DALL-E generated images for the same labels. Right: sample of MiniGPT-v2 responses given TSI  and DALL-E images indicated by a yellow arrow.}
\vspace{-0.5cm}
\end{figure}

Despite the effectiveness of vision encoder updates, continuous and frequent updates can lead to performance deterioration. Therefore, we recognize the need for not only efficient updates but also localization to the data at hand, in order to limit degradation in unrelated areas of knowledge. Parameter-efficient fine-tuning methods often degrade unrelated knowledge similar to full fine-tuning, as shown previously by \cite{zhang2023overcoming} and as demonstrated in our experiments. For instance, LoRa’s low-rank updates still alter all model parameters, resulting in performance decline upon multiple updates.

To achieve localized updates, we propose that only parameters relevant to the task at hand should be modified while keeping the rest of the model’s parameters intact. A similar direction is followed in Language Model Editing, although existing approaches are usually specific to factual knowledge updates and cannot be easily applied to vision-related updates.

In order to identify which parameters should be updated, we construct the selection process as masking of parameters on which said masks should preserve the gradient norm of the estimated update of the model. This can be achieved by selecting parameters with the greatest gradient norm. Now, for MLP layers of the transformer model, we first follow a similar approach to SPU~\cite{zhang2023overcoming} by selecting the top $k$ parameters according to the gradient norm. Our approach is generalizable to attention heads, where we select specific attention heads based on the same rule. We  combine the estimated masks with low rank updates~\cite{hu2021lora}, in a unique way, achieving both locality and efficiency. %Inspired by the Selective and Localized Updates method (SPU), we propose a generalized approach that updates both attention heads and MLP layers in a transformer model. Our method combines SPU with LoRa for increased efficiency. Initially, we identify which parameters in the MLP layer and which attention heads are specialized and relevant to the specific data. We then update only the selected parameters and attention heads using LoRA. Our selection criterion, based on gradient scores, approximates the parameters that yield the most significant loss reduction with minimal changes as shown by our analysis. 

We validate our method across various benchmarks, both by updating CLIP and by enhancing VLM models based on CLIP. Our approach demonstrates superior performance and preserves the model’s generic knowledge. While our focus lies on updating the vision encoder, our method is generic and applicable to any transformer model whether for vision, language, or any other modality.
Our contribution are as follows: 
1) We evaluated CLIP on out-of-distribution benchmarks and observed shortcomings in certain scenarios. These limitations are then propagated to the VLMs that leverage CLIP’s embeddings.
2) Our study demonstrates that updating the vision encoder separately, specifically on data where CLIP fails, can significantly correct VLM mistakes on previously unseen images from this data. %Importantly, this approach proves to be more robust against catastrophic forgetting of the model’s generic knowledge compared to updating the language model.
3) We propose a novel parameter-efficient tuning method that not only targets efficiency but also ensures the preservation of the model’s generic knowledge. Our method exhibits robustness in both few-shot learning and continual updates, achieving new state-of-the-art results.

\section{Is CLIP truly robust to all conditions?}\label{sec:clip-robustness}
VLMs are either trained in an end-to-end fashion or made as a composite of separate vision and language models. In the latter, it is typically the vision encoder that remains frozen throughout the training process of the VLM. CLIP~\cite{radford2021learning}, a vision transformer model, trained by contrasting vision and language, is the primary vision encoder deployed in this line of VLMs. For instance, it is used in such fashion in MiniGPT, MiniGPTv2, BLIP2~\cite{li2023blip}, CogVLM~\cite{wang2023cogvlm}, Kosmos-2~\cite{peng2023kosmos}, LLaVA~\cite{liu2024visual} and LLavaNext~\cite{liu2024llava}.
CLIP has been trained on large numbers of vision and language pairs. When testing CLIP models in challenging domains, including adversarial scenarios~\cite{mayilvahanan2023does}, the model demonstrates unprecedented robustness to various domain shifts. It was also noted by ~\cite{mayilvahanan2023does} that the large pretraining dataset of CLIP might contain representative examples of those out of distribution benchmarks, which raises questions about the real robustness of CLIP.
%   \end{minipage}\hfill
%   % \resizebox{.37\textwidth}{!}{
%   \begin{minipage}[b]{.35\linewidth}
%     \centering
%       \begin{minipage}[b]{1\linewidth} 
%   \resizebox{1\textwidth}{!}{
% \begin{tabular}{l c c c c c c}
% \toprule
%    &  \multicolumn{3}{c}{\textbf{EVA-Clip-G-14}} & \multicolumn{3}{c}{\textbf{ OpenAI-Clip-L-14}}  \\
%   \cmidrule(lr){2-4} \cmidrule(lr){5-7} \textbf{Dataset} & \textbf{ImageNet} & \textbf{TSI} & \textbf{DALL-E}& \textbf{ImageNet} & \textbf{TSI} & \textbf{DALL-E}\\
% \midrule
% \textbf{Zero-Shot} &  $78.5$ & $11.4$ &$97.0$& $76.6$ & $13.2$ &$90.0$\\
% \bottomrule
% \end{tabular}}
%     \captionof{table}{{\label{CLIP_TSI}\small CLIP EVA-ViT-g-14 and OpenAI-Clip-L-14 Accuracy on TSI dataset.} }
%   \end{minipage}
%   \vfill
%      \begin{minipage}[b]{1\linewidth}
%     \centering
%    %  \resizebox{.15\textheight}{!}{
%    \includegraphics[width=1\linewidth]{plots/TSI-MiniGPT.pdf}
% %   }
% \captionof{figure}{\label{fig-Dalle-tsi}\small }
%   \end{minipage}
% %}
%   \end{minipage}%}

% \begin{figure*}[t]
% \centering
% \includegraphics[width=0.6\linewidth]{plots/TSI-Dalle.pdf}
% \caption{Comparison of images from TSI dataset compared to DallE generated images for the same labels.}
% \end{figure*}
\begin{figure}[t]
  \begin{minipage}[b]{0.42\linewidth} 
   \resizebox{1\textwidth}{!}{
      \begin{tabular}{ c c c}
%\footnotesize
\toprule
Model  &{DALL-E-VQA}  & {TSI-VQA} \\
\midrule
Llama2+ln& $86.5$ & $82.3$\\
EVA-CLIP-G &  $\mathbf{87.1}$&$\mathbf{86.0}$ \\
\bottomrule
\end{tabular}}
 \captionof{table}{\label{table:fine_tune_llm_vs_vit} \footnotesize  MiniGPTv2 VQA Accuracy (\%) after finetuning the LLM (with LoRa $r=64$) compared to finetuning EVA-CLIP-G \textit{separately}.}
  \end{minipage}
  \hfill
\begin{minipage}[b]{0.54\linewidth} 
  \resizebox{1\textwidth}{!}{
\begin{tabular}{c c c c c c}
\toprule
    \multicolumn{3}{c}{\textbf{EVA-Clip-G-14}} & \multicolumn{3}{c}{\textbf{ OpenAI-Clip-L-14}}  \\
  \cmidrule(lr){1-3} \cmidrule(lr){4-6}  
  \textbf{ImageNet} & \textbf{TSI} & \textbf{DALL·E}&
  \textbf{ImageNet} & \textbf{TSI} & \textbf{DALL·E}\\
\midrule
 $78.5$ & $11.4$ &$97.0$& $76.6$ & $13.2$ &$90.0$\\
\bottomrule
\end{tabular}}
    \captionof{table}{{\label{CLIP_TSI}\footnotesize CLIP EVA-ViT-G-14 and OpenAI-Clip-L-14 zeroshot Accuracy (\%) on ImageNet, TSI and DALL·E datasets. TSI accuracy is much lower than DALL·E.} }
  \end{minipage}
  %\vspace*{-1cm}
  \end{figure}
We investigate the out-of-distribution generalization capabilities of pretrained vision encoders, in order to assess whether future model updates are indeed unnecessary or whether simply most publicly available datasets used for testing are indeed represented in the pretraining data. We first considered the XImageNet-12 benchmark~\cite{li2023ximagenet}, a recent benchmark designed to examine the robustness of vision models, by simulating diverse out of distribution effects. The basic CLIP model, CLIP-ViT-B-16, shows performance impressively close to perfect on most domains. Our results and a full description of the dataset can be found in Appendix~\ref{sec:ximagenet}.

% Although only one domain with random backgrounds of other objects exhibits weak performance, this could be attributed to model confusion between the two objects in the foreground and background, rather than a weakness in understanding the image.
However, XImageNet-12 is generated based on ImageNet images, and hence the generalization to never seen images  under distribution shift is still not proven.
%Taken into account that  images of XImageNet-12 are generated by models that leverage CLIP embedding, could explain the strong performance of CLIP on those domain shifts. %This could be a reason why CLIP understands these distribution shifts. 
% Further, as noted by~\cite{mayilvahanan2023does} the large web scale dataset used for training CLIP make us question whether CLIP has already seen similar examples to all those benchmarks designated for out of distribution robustness. 
We then conducted a simple yet realistic evaluation. We considered Toyota Smart Home (TSI) dataset~\cite{das2019toyota} a dataset of daily living activities staged in a home-like environment.
This dataset cannot be publicly crawled from the web, it is only accessible upon request. This makes it unlikely to have been used to train CLIP. Further, the data depict elderly people activities (age bias), blurred faces (blurring effect) and is captured from a mounted camera with somewhat low resolution, yet the actions are easily recognizable to the human eye. For more details, refer to the Appendix and Experiments Section~\ref{sec:experiments}.

Surprisingly, when evaluating CLIP on images from the TSI dataset, we observed only moderate performance and encountered numerous mistakes. Table~\ref{CLIP_TSI} reports CLIP accuracy on TSI dataset compared to ImageNet.
Now, we consider whether a VLM using CLIP's vision encoder would be able to describe the activity in these images accurately or not.
Further testing of a VLM (MiniGPTv2~\cite{chen2023minigpt}) using CLIP as the vision encoder revealed similarly poor performance. We refer to Figure~\ref{fig-Dalle-tsi} for an example response and to the Appendix for more qualitative results. The main failure modes present are hallucination of wrong activities or describing the background rather than the action.
This suboptimal performance could be attributed to either CLIP’s limited knowledge of the performed activities or its lack of robustness to the distribution shift present in the images.
To address the first assumption, we leveraged diffusion models, specifically DALL·E 2, to generate images of people performing the same actions. After verifying the accuracy of these generated images, we tested CLIP’s predictions on them. Remarkably, CLIP accurately recognized the actions on the synthetic images. Similarly,  MiniGPTv2~\cite{chen2023minigpt} also provided very accurate descriptions of the generated images. Figure~\ref{fig-Dalle-tsi} show some generated images compared to images from TSI~\cite{das2019toyota} of similar activities.

Finally, we conducted an experiment where we finetune CLIP encoder on a training split of TSI dataset. We then plug CLIP back into MiniGPTv2 and evaluate VQA accuracy on TSI and DALL-E datasets. For a sanity check, we also finetune the LLM decoder (Llama2) and the linear layer using LoRa~\cite{hu2021lora}. Results are shown in Table~\ref{table:fine_tune_llm_vs_vit}. 
Finetuning the vision encoder separately can drastically improve the performance of MiniGPTv2 on both TSI and DALL-E images, with $4\%$ improvement over finetuning the LLM.  Note that optimizing the LLM needs careful attention and hyper-parameter tuning, in contrast, tuning the CLIP model is comparatively straight-forward. Given the difference in resources needed to update CLIP as opposed to updating the LLM, we suggest that updating the vision encoder can be an efficient alternative to fully updating the LLM. 
% When encountering new domains, 

% \begin{wraptable}{r}{4cm}
% \resizebox{0.3\textwidth}{!}{
% \begin{tabular}{ c c c}
% %\footnotesize
% \toprule
% Model  &{DALL-E-VQA}  & {TSI-VQA} \\
% \midrule
% Llama2& $86.5$ & $82.3$\\
% CLIP-G &  $\mathbf{87.1}$&$\mathbf{86.0}$ \\
% \bottomrule
% \end{tabular}}
% \caption{\label{table:fine_tune_llm_vs_vit} \footnotesize Accuracy (\%) scores for MiniGPTv2 when finetuning Llama2 (with LoRa $r=64$) and when full finetuning EVA-CLIP-G \textit{separately}.} 
% \vspace{-2cm}
% \end{wraptable}
% This empirical evidence highlights the imperfections of the CLIP model and underscores the need for continuous improvements when encountering new domains and out-of-distribution data. It further suggests that an efficient robust method for updating CLIP can be a powerful alternative to finetuning the large language model. 
%For qualitative results comparing the generated images to those from the TSI dataset, please refer to the accompanying figure.
\section{\texorpdfstring{\underline{Lo}w-\underline{R}ank Adaptation with \underline{S}tructured \underline{U}pdates}{Lg}}
\label{sec:lorsu_method}
To efficiently fine-tune the large visual encoder\footnote{The methodology can be easily applied to LLMs as well since the method is designed for transformer-based architectures.} without incurring catastrophic forgetting, we develop a new parameter-efficient fine-tuning method, called Low-Rank Adaptation with Structured Updates (\textbf{LorSU}). LoRSU aims to update specific parameters from each transformer block in a parameter efficient way while circumventing the risk of generic knowledge loss due to fine-tuning on the current task. To achieve this, we update a small part of the parameters of the first linear layer of the MLP block in the transformer module of each layer as in \cite{zhang2023overcoming}. However, this approach might lack flexibility since all the other parameters from the transformer block remain frozen during optimization. To overcome this, we also update the most informative attention heads based on the gradient information of the loss function.

More specifically, let a dataset $\cD_t = \{\bx_n, \by_n \}_{n=1}^{N_t}$ for the current task $t$ where $\bx_n$ is an image with text description $\by_n$ and $\cL(\btheta; \cD_t) := \cL_t(\btheta)$ is the contrastive loss used for CLIP pretraining and $\btheta \in \mathbb{R}^d$ is the full set of model's parameters. The standard Multi-head Self-Attention Mechanism~\cite{vaswani2017attention}, comprised of $H$ heads, is defined as the concatenation of multiple self-attention (SA) blocks:
\begin{align}
        \bq^{(i)} & = W_q^{(i)} Z^{\top}, \bk^{(i)} = W_k^{(i)} Z^{\top}, \bv^{(i)} = W_v^{(i)} Z^{\top} \in \mathbb{R}^{D_h \times N}, \\
        A^{(i)} & = \text{softmax}( \bq^{(i)^\top} \bk^{(i)} / \sqrt{D_h} )  \in \mathbb{R}^{N \times N}, \\
        \text{SA}_i(Z) & = A^{(i)} \bv^{(i)^\top}  \in \mathbb{R}^{N \times D_h}, ~~i=1, \ldots, H.
\end{align}
where $Z \in \mathbb{R}^{N \times D}$ is the input matrix of $N$ tokens of dimension $D$ and $ W_q^{(i)},  W_k^{(i)},$ and $ W_k^{(i)}$ are the query, key, and value matrices of learnable parameters for head $i$, respectively. The final MSA function is defined as 
\begin{align}
        \text{MSA}(Z) = \text{Concat}\left[ SA_1(Z), \ldots, SA_H(Z)  \right] W_o \in \mathbb{R}^{N \times D},~~~~~ W_o \in \mathbb{R}^{H D_h \times D}, \label{eq:msa}
    \end{align}
Since we care to update the parameters of the heads that cause the largest changes in  $\cL_t(\btheta)$, we compute the gradient of the loss with respect to the parameters of each head and then we pick to update those heads that their cumulative contribution to the loss change is the largest. Since the matrices $W_q^{(i)},  W_k^{(i)},  W_v^{(i)}$ are all the parameters of head $i$, we can define an importance score for each head by adding the squared values of their corresponding gradients $G_q^{(i)} = \nabla_{W_q^{(i)}} \mathcal{L}$, $G_k^{(i)} = \nabla_{W_q^{(i)}} \mathcal{L}$, $G_v^{(i)} = \nabla_{W_v^{(i)}} \mathcal{L}$, and $G_o^{(i)} = \nabla_{\widetilde{W}_o^{(i)}} \mathcal{L}$, i.e.
\begin{equation}
    s_i = \sum_{m,l} \left( (G_q^{(i)}[m,l])^2 + (G_k^{(i)}[m,l])^2 + (G_v^{(i)}[m,l])^2 + (G_o^{(i)}[m,l])^2 \right). \label{eq:s_i} 
\end{equation}
We provide a theoretical justification of \eqref{eq:s_i} in the next section. We choose the parameters of the top-k heads $\{s_1, \ldots, s_H \}$, $I \subset \{1, \ldots, H \}$, to be updated on the current task. However, the number of parameters remain high due to the large weight matrices. Therefore, we opt for parametrizing the original weights using LoRA parametrization~\cite{hu2021lora} to reduce computational burden. The matrices $W_q^{(i)}, W_k^{(i)}, W_v^{(i)}, i \in I$ are now defined as
\begin{align}
    W_q^{(i)^{\prime}} & = W_q^{(i)} + A_q^{(i)} B_q^{(i)} \\
    W_k^{(i)^{\prime}} & = W_k^{(i)} + A_k^{(i)} B_k^{(i)} \\
    W_v^{(i)^{\prime}} & = W_v^{(i)} + A_v^{(i)} B_v^{(i)}.
\end{align}
Finally, to ensure that we only update $W_q^{(i)},  W_k^{(i)},  W_v^{(i)}, \forall i \in I$ we use a binary mask on the gradient vector with respect to all parameters of all attention heads. We keep the projection matrix $W_o$ frozen throughout optimization.

Regarding the first linear layer in the MLP module, $W_{\text{fc1}} \in \mathbb{R}^{d \times D}$, we mask the gradients of $W_{\text{fc1}}$ so only the most important parameters for the current task to be updated, i.e. we use the following biased gradient update.
\begin{equation}
\hat{\nabla}_{W_{\text{fc1}}} \cL_t = M_{\text{fc1}} \odot \nabla_{W_{\text{fc1}}} \cL_t,
\end{equation}
where  $M_{\text{fc1}} \in \{0, 1 \}^{d \times D}$ is a zero-one mask that is built by choosing a proportion of the largest squared values of $\nabla_{W_{\text{fc1}}} \cL_t$ in a similar manner as in~\cite{zhang2023overcoming} and $\odot$ is the Hadamard product.

\subsection{Theoretical justification}
The scores in \eqref{eq:s_i} can be derived from the following constrained (binary) optimization problem
\begin{align}
  \bp^*  =  \argmax_{\bp \in \{0, 1 \}^d} \frac{\norm{\bp \odot \nabla_W \cL(\btheta_0) }^2}{\norm{\nabla_W \cL(\btheta_0)}^2},  \label{eq:maxim} 
 ~~~~& \text{s.t.}~~~ \bigcup_{\ell=1}^G I_{\ell} \subset \{1, 2, \ldots, d\},~~\text{where}~~ I_i \cap I_j = \emptyset,~~\forall i\neq j, \nonumber \\
  & S = \sum_{\ell=1}^G s_{\ell},~~s_{\ell} \leq |I_{\ell}|~~\forall \ell, ~~~\norm{\bp}_0 \leq S, 
\end{align}
Here $\btheta_0$ is the pretrained vector of parameters before we use the $\cD_t$ for fine-tuning. The mask $\bp^*$ is chosen so that the gradient norm is preserved against masking as much as possible under these constraints.
\begin{definition}\label{def:tops}
The operator TOP-$S: \mathbb{R}^d \rightarrow \mathbb{R}^d$, for $1 \leq S \leq d$ is defined as
\begin{equation}\
    \left( \text{TOP-}S(\bx) \right)_{\pi(i)} := \left\{
\begin{array}{ll}
      x_{\pi(i)}, & i \leq S \\
      0, & \text{otherwise},\\
\end{array} 
\right. \nonumber
\end{equation}
where $\pi$ is a permutation of $\{1, 2, \ldots, d \}$ such that $|x_{\pi(i)}| \geq |x_{\pi(i+1)}|$, for $i=1, \ldots, d-1$, i.e. the TOP-$S$ operator keeps only the $S$ largest elements of $\bx$ in magnitude and truncates the rest to zero.
\end{definition}

\begin{lemma}\label{lemma:max_norm}
For any $\bx \in \mathbb{R}^d-\{ \mathbf{0}\}$, $1 \leq S \leq d$, the optimal mask
\begin{align}
     \bp^* & =  \argmax_{\bp \in \{0, 1 \}^d} \frac{\norm{\bp \odot \bx }^2}{\norm{\bx}^2},~~\text{s.t.}~\norm{\bp}_0 \leq S,  \nonumber
\end{align}
has zeros everywhere except the $S$ largest elements of $\bx$ in magnitude.
\begin{proof}
    Rewriting the optimization problem as 
    \begin{equation}
        \max_{\bp \in \{0, 1 \}^d} \sum_{i=1}^d p_i x_i^2,~~ \text{s.t.}~\sum_{i=1}^d p_i \leq S \nonumber,
    \end{equation}
    we notice that this a trivial binary knapsack problem with maximum weight capacity $S$ and weights equal to one. Hence, the maximum is attained when we pick the $S$ maximum $x_i^2$. 
\end{proof}
\end{lemma}

\begin{remark}

\end{remark} It holds that $\text{TOP-}S(\bx) = \bp^* \odot \bx$.

\begin{corollary}
The optimal mask $\bp^*$ in \eqref{eq:maxim} has zeros everywhere except for the indices $i \in \{j: \exists \ell \in \{1, \ldots, G \},~\text{such that}~j \in \{ \pi_{\ell}(1), \ldots,  \pi_{\ell}(s_{\ell})\}    \}$, where $\pi_{\ell}$ is the same permutation as in Definition $\ref{def:tops}$ for the set of indices $I_{\ell}
$.
\end{corollary}
\begin{proof}
    The result follows from the mutual exclusiveness of $I_{\ell}$ in the constraints of \eqref{eq:maxim} and Lemma \ref{lemma:max_norm}.
\end{proof}

%\begin{assumption}\label{assumption:beta}
%It holds that $\forall~1 \leq S \leq d,~\exists~\beta \in [0,1)$ such that the optimal mask $\bp^*$ satisfies the following inequality: 
%\begin{equation}
%    \frac{\frac{d-S}{d}\norm{\bp^* \odot \nabla\cL(W) }^2 + \norm{(\mathbf{1} - \bp^*) \odot \nabla\cL(W) }^2}{\norm{ \nabla\cL(W)}^2} \leq \beta ,~~\forall W \in \mathbb{R}^d.
%\end{equation}
%\end{assumption}

%\begin{assumption}[$\beta$-bounded bias]\label{lemma:bounded_bias}
%For the optimal mask $\bp^*$, it holds that $\forall~1 \leq S \leq d,~\exists~ 1 \leq \beta \leq 1 + \frac{S}{d}$ such that 
%\begin{equation}
%    \norm{(\mathbf{1} - \bp^*) \odot \nabla\cL(W)}^2 \leq \frac{\beta (d - S)}{d} \norm{ \nabla\cL(W)}^2,~~\forall W \in \mathbb{R}^d.
%\end{equation}
%\end{assumption}
\section{Related Work}\label{sec:related_work}

%\paragraph{Model Editing}
Large language model and vision language models are strong foundation models but are still prone to mistakes and their knowledge can get outdated, consequently it is important to develop efficient updates that preserve unrelated knowledge.  
The main line of work in this area focuses on LLM editing, where previous factual knowledge has changed and the model must be updated effectively on those changes. 
Most notably \cite{meng2022locating} and \cite{ilharco2022editing} first analyze the models to identify specific layers for editing, i.e., where factual knowledge are "stored" in the model, and then apply algebra-based or meta-learning methods to adjust the weights of these localized layers. To insure the locality of the updates these methods usually leverage additional sets representing unrelated factual knowledge.

Another line of work focus on updating the model for a new task or dataset with parameter efficient finetuning. Low Rank Updates (LoRA)~\cite{hu2021lora} approximates the parameter updates by a low rank matrix, achieving similar performance on the target task by optimizing only 1\% of the parameters compared to the full model. The original version of LoRA updated only attention layers. Subsequently, several extensions have been proposed to enhance LoRA which modify all layers. Various options are available including  adapting the learning rate~\cite{hayou2024lora+} of the low rank matrix, using an adaptive rank~\cite{zhang2023adaptive} or decomposing the update matrix into magnitude and direction~\cite{liu2024dora}. These approaches focus solely on efficiently updating the network without considering the impact on model performance for other unrelated tasks or enforcing any locality to specific layers or parameters. It is worth noting that LoRA drop~\cite{zhou2024lora} attempts to localize the updates to specific layers.  It initially allows a few iterations of LoRA updates and then assesses the impact of each low-rank update on individual layers and selectively updates only those layers where the change exceeds a specified threshold. However, this selectivity remains at the layer level and depends on the change introduced by a few full updates.
In contrast,  we treat each layer differently based on its structure and assess the relevance of individual parameters to the task at hand. We then holistically combine the importance and relevance of these parameters with low-rank updates.
 
In the context of updating vision models for specific tasks, SPT~\cite{he2023sensitivity} estimates a mask of updates based on parameter sensitivity to the task. Depending on the number of relevant parameters, either low-rank or sparse updates are performed (using a threshold).
With regards to continual updating CLIP while maintaining its generalization performance and reducing forgetting, SPU~\cite{zhang2023overcoming} treats layers of the transformers differently, and inspired by knowledge neuron theory, SPU localizes the updates to the first feedforward layer of each transformer block and then only relevant parameters to the task at hand are updated.  We further refer to~\cite{de2021continual} for a survey on continual learning.
In our approach, we select and identify relevant parameters to the current data. However, we generalize the updates to all layers while preserving  the specificity of each layer. We choose masks that maintain the gradient norm of parameter updates and combine them with LoRA on selected attention heads, striking a balance between  adaptivity and stability.

\section{Experiments}\label{sec:experiments}
The section serves to support our conclusions on CLIP's robustness and how finetuning it separately, with our method, can result in a robust performance both in offline and continual few-shot updates. The robust performance after updates is reflected in evaluations using two VLMs.
\subsection{Setting}
In the following we describe the construction of TSI based datasets, and other used datasets, training protocols,   models considered, and methods compared.\\
%\paragraph{Datasets}
\myparagraph{Classification datasets.}
\textbf{TSI}: We process the TSI~\cite{das2019toyota} dataset as an image classification dataset where the target is to recognize the activity depicted in each image.  We extract frames from videos and create a train set of approximately 10K images and a test set of approximately 5k images. We consider 22 represented classes of activities. 
\textbf{DA·E}: We consider the same 22 classes of activities represented in TSI and query DALL·E 2 to generate representative images of these activities. We extract 30 images per action totaling 660, all of them are designated for testing. 
\textbf{ImgNet}: We consider ImageNet~\cite{imagenet} as a control set to measure how much CLIP models' performance deteriorates after being tuned on other datasets.
\textbf{GTS}~\cite{stallkamp2012man} the German Traffic Sign dataset. In \cite{zhang2023overcoming}  GTS was considered as out of distribution for CLIP pretraining, CLIP zero shot performance  is significantly lower than the performance of a linear classifier trained on ResNet50 features~\cite{radford2021learning}.

\myparagraph{Visual Question Answering datasets.}
To evaluate how the examined VLM is performing before and after the vision encoder update, we consider 3 visual question answering datasets: 
\textbf{VizWiz}~\cite{gurari2018vizwiz}: a visual question answering dataset designed to mimic the scenario where a visually impaired person takes a photo by phone and asks the visual assistant a question about it. It consists of about 31,000 visual questions.
\textbf{VSR}~\cite{liu2023visual}:  a dataset for Visual Spatial Reasoning with more than 10k
natural text-image pairs and 66 types of spatial relations.
HM~\cite{kiela2020hateful} hateful memes dataset designed to detect multimodal hateful memes.
\textbf{TSI}: We convert the TSI classification dataset into a VQA dataset with multiple choice responses and measure the likelihood of the correct response following common practice in VQA evaluation. 
\textbf{DA·E}: Following the practice for TSI, we convert DA·E generated dataset to VQA format.

\myparagraph{Training protocols.}
\textbf{Offline}: We first consider an offline fine-tuning setting as a sanity check where the vision encoder is updated offline on the full training set of TSI data.  The goal is to asses the performance of CLIP before and after the update by different methods and the VLM responses when the updated vision backbone is plugged in. \\
\textbf{Continual \& few-shot}: We design this setting to imitate a realistic scenario where the model is updated on images where it makes mistakes with few-shot examples, and the process is to be repeated as long mistakes are shown. We follow the common practice in continual fewshot learning~\cite{panos2023first}  to construct the sequences. We divide the dataset into 5 sets of disjoint classes and consider 20 shot setting where only 20 training examples of each action is provided. Accuracy is measured on the full test set. In the Appendix we consider 50 shots setting. We always report the accuracy (classification or VQA) on the test sets of the concerned datasets at the end of a training sequence, with that we measure the ability of the model to accumulate knowledge and resist forgetting~\cite{de2021continual}. Note that we do not consider any replay~\cite{er} of samples from classes of previous sessions. 

\textbf{Implementation details.} We refer to the Appendix~\ref{sec:implementation} for implementation details.

\myparagraph{Models.} We consider two versions of CLIP, namely OpenAI-CLIP-L-14 and EVA-CLIP-G-14. OpenAI-CLIP-L-14~\cite{radford2021learning}  is a large ViT pretrained by OpenAI.  EVA-CLIP-G-14, a giant ViT, is  an improved version with some set of optimization and augmentation techniques~\cite{sun2023eva} pretrained on LAION-400M dataset~\cite{schuhmann2021laion}.
For vision language models, we consider two popular vision language models that leverage a frozen CLIP image encoder. 
LLaVA~\cite{liu2024visual} Large Language and Vision Assistant, connects pretrained and frozen OpenAI-CLIP-L-14 with a LLM (Vicuna-7b~\cite{chiang2023vicuna}) though a linear layer. The LLM and the linear layer are optimized during the visual instruction tuning process while CLIP remains frozen. %optimizes a pretrained LLM and a linear  to generate multimodal language-image
%Instruction-following data
MiniGPTv2~\cite{chen2023minigpt} concatenates adjacent tokens from EVA-CLIP-G-14. OpenAI image embedding and process it with a linear layer as input to LLama-2 (7B-chat)~\cite{touvron2023llama}. Similar to   LLaVA~\cite{liu2024visual} the linear layer and the language model are optimized while the vision encoder remains frozen.

\myparagraph{Methods.}
When finetuning CLIP, we fine-tune both visual and text encoders following~\cite{goyal2023finetune} with the same contrastive image language loss used in the pretraining of CLIP. 
We consider the following methods for fine-tuning.
\textbf{F-FT}: Full fine-tuning of all model parameters. This can provide the best accuracy, but is prone to forgetting and overfitting.
\textbf{LN}: Optimization of the layer norm parameters of the transformer, an adaptation of~\cite{shysheya2022fit}. This approach modifies a very small fraction of parameters and has been shown to be a robust approach for few-shot updates \cite{panos2023first}. %\rahaf{Any reference we can follow?} \ares{Maybe this one~\cite{shysheya2022fit}? We could also cite our paper here \cite{panos2023first} since we are on a CL setting and we learn the layer norm paramaters (aka FiLM parameters)}
\textbf{LoRA}: we consider Low rank updates~\cite{hu2021lora} applied to all transformer layers.
\textbf{SPU}: Selective Parameters Updates~\cite{zhang2023overcoming} is a recent method proposed to continually update CLIP with minimal forgetting and generic knowledge loss. 
\textbf{\ours:} We coin our approach \ours as short for Low Rank Structured Updates.
We always report the zero-shot performance of the model (without training), we refer to this as \textbf{Zr-shot}.
\subsection{Results}
\begin{table}[t]
%\vspace*{-1cm}
\centering
\small
  \resizebox{1\textwidth}{!}{
   \begin{tabular}{l c c c c c c| c c c c c c}
\toprule
&  \multicolumn{6}{c|}{\textbf{TSI}} &  \multicolumn{6}{c}{\textbf{GTS}} \\
\cmidrule(lr){2-7} \cmidrule(lr){8-13} 
   &  \multicolumn{3}{c}{\textbf{EVA-Clip-G-14}} & \multicolumn{3}{c|}{\textbf{ OpenAI-Clip-L-14}} &  \multicolumn{3}{c}{\textbf{EVA-Clip-G-14}} & \multicolumn{3}{c}{\textbf{ OpenAI-Clip-L-14}} \\
  \cmidrule(lr){2-4} \cmidrule(lr){5-7} \cmidrule(lr){8-10} \cmidrule(lr){11-13}  \textbf{Dataset} & \textbf{ImgNet} & \textbf{DA·E} & \textbf{TSI}  & \textbf{ImgNet} & \textbf{DA·E} & \textbf{TSI} &  \textbf{ImgNet} & \textbf{DA·E} & \textbf{GTS}  & \textbf{ImgNet} & \textbf{DA·E} & \textbf{GTS} \\
\midrule
\textbf{Zr-Shot} & $78.5$ & $97.0$ & $11.4$ & $76.6$ & $90.9$ & $13.2$&$78.5$ & $97.0$ & $49.1$ & $76.6$ & $90.9$ & $52.4$  \\
\textbf{F-FT} & $74.6$ & $88.6$ & $\mathbf{77.7}$ & $73.9$ & $88.3$ & $\mathbf{80.9}$&$73.2$ & $94.7$ & $\mathbf{99.6}$ & $71.1$ & $88.2$ & $\mathbf{99.3}$  \\
%\midrule
\textbf{LN} & $65.6$ & $96.4$ & $68.1$ & $64.7$ & $81.4$ & $73.8$&$70.7$ & $93.3$ & $97.6$ & $52.0$ & $79.2$ & $97.9$ \\
%\midrule
%\midrule
\textbf{LoRA} & $75.8$ & $92.4$ & $75.6$ & $72.8$ & $88.1$ & $79.2$& $75.4$ & $94.9$ & $99.1$ & $71.7$ & $89.1$ & $99.0$  \\
%\midrule
\textbf{SPU} & ${78.4}$ & ${96.5}$ & $72.8$ & ${76.1}$ & $88.9$ & $75.3$& ${78.4}$ & ${96.4}$ & $98.6$ & $76.5$ & ${90.9}$ & $98.4$  \\
%\midrule
\textbf{LoRSU} & $78.1$ & $95.6$ &  \underline{$77.1$} & $76.0$ & ${91.8}$ & \underline{$79.6$}& ${78.4}$ & $96.2$ &  \underline{$99.5$} & $76.2$ & $90.8$ & \underline{ ${99.2}$} \\
\bottomrule
\\
\end{tabular}}
\caption{\footnotesize\label{table:clip_acc_tsi_offline}Accuracy (\%) scores for CLIP EVA-ViT-G-14 and OpenAI-ViT-L-14 models for various fine-tuning methods. All use the whole training split (\textbf{offline}) of \textbf{TSI}~(left) and  \textbf{GTS}~(right) to fine-tune the visual encoder, best method on the target dataset is highlighted in bold,  second best method is underscored. Ours,  \ours, achieves  very close accuracy to  fully finetuning (F-FT) dataset with negligible deterioration on other datasets.}
%\vspace*{-1cm}
\end{table}
\begin{table}
\centering
\small
  \resizebox{1\textwidth}{!}{
   \begin{tabular}{l c c c c c c| c c c c c c}
\toprule
&  \multicolumn{6}{c|}{\textbf{TSI}} &  \multicolumn{6}{c}{\textbf{GTS}} \\
\cmidrule(lr){2-7} \cmidrule(lr){8-13} 
   &  \multicolumn{3}{c}{\textbf{EVA-Clip-G-14}} & \multicolumn{3}{c|}{\textbf{ OpenAI-Clip-L-14}} &  \multicolumn{3}{c}{\textbf{EVA-Clip-G-14}} & \multicolumn{3}{c}{\textbf{ OpenAI-Clip-L-14}} \\
  \cmidrule(lr){2-4} \cmidrule(lr){5-7} \cmidrule(lr){8-10} \cmidrule(lr){11-13}  \textbf{Dataset} & \textbf{ImgNet} & \textbf{DA·E} & \textbf{TSI}  & \textbf{ImgNet} & \textbf{DA·E} & \textbf{TSI} &  \textbf{ImgNet} & \textbf{DA·E} & \textbf{GTS}  & \textbf{ImgNet} & \textbf{DA·E} & \textbf{GTS} \\
\midrule
\textbf{Zr-Shot} & $78.5$ & $97.0$ & $11.4$ & $76.6$ & $90.9$ & $13.2$ & $78.5$ & $97.0$ & $49.1$ & $76.6$ & $90.9$ & $52.4$\\
%\midrule
\textbf{F-FT} & $62.5$ & $77.4$ & $37.0$ & $68.8$ & $81.7$ & $40.8$ & $78.2$ & $96.8$ & $46.5$ & $75.1$ & $89.6$ & $38.0$\\
\textbf{LN} & $27.8$ & $72.6$ & $40.9$ & $28.3$ & $60.2$ & $45.3$ & $78.0$ & $96.8$ & ${48.1}$ & $75.5$ & $90.6$ & $49.1$  \\
%\midrule
%\midrule
\textbf{LoRA} & $57.1$ & $67.3$ & $37.9$ & $48.2$ & $52.9$ & $41.4$  & $76.5$ & $95.5$ & $35.8$ & $67.3$ & $83.5$ & $35.1$ \\
%\midrule
\textbf{SPU} & $76.7$ & $85.3$ &\underline{$44.0$} & $74.2$ & $85.2$ & \underline{$48.8$} & $78.5$ & $96.8$ & \underline{$54.1$} & $76.6$ & $90.5$ & \underline{$54.4$} \\
%\midrule
\textbf{LoRSU} & $78.1$ & $95.6$ & $\mathbf{48.7}$ & $76.1$ & $91.1$ & $\mathbf{51.6}$ & $78.5$ & $96.8$ & {$\mathbf{56.6}$}& $76.6$ & $90.9$ & $\mathbf{57.3}$  \\
\bottomrule
\end{tabular}}
\caption{\footnotesize\label{table:clip_acc_tsi_fs20_cl}\textbf{Last session} accuracy (\%)  scores for CLIP EVA-ViT-G-14 and OpenAI-ViT-L-14  models for various fine-tuning methods. The model is fine-tuned on each session (5 in total) of \textbf{TSI} (left) and \textbf{GTS} (right)  using \textbf{20 shots} per session, best method on the target dataset is highlighted in bold,  second best method is underscored. Our  \ours achieves the best performance on the target dataset with minimal deterioration on other datasets.}
\vspace{-0.1cm}
\end{table}
% Accuracy (\%) scores for CLIP EVA-ViT-G-14 and OpenAI-ViT-L-14 models for various fine-tuning methods. All use the whole training split (\textbf{offline}) of \textbf{TSI}~(left) and GTS~(right) to fine-tune the visual encoder, best method on the target dataset is highlighted in bold,  second best method is underscored. Ours,  \ours, achieves  very close accuracy to  fully finetuning (F-FT) dataset with negligible deterioration on other datasets.
\paragraph{Offline updates.}
We first consider the common finetuning scenario where we fully finetune the CLIP encoder on the considered dataset and then evaluate the performance on classification and VQA tasks. Table~\ref{table:clip_acc_tsi_offline} reports the accuracy after finetuning CLIP models on TSI and  GTS datasets (in separate sessions). On the target dataset, F-FT achieves the best accuracy  but lacks behind other methods on  ImgNet and even on DA·E. For example, using an EVA-CLIP backbone in the TSI dataset, F-FT accuracy decreases on DA·E from 97\% to 88.6\% after fine-tuning. Our method \ours achieves a performance close to that of F-FT on the target dataset while keeping the accuracy on ImgNet and DA·E almost unchanged. LoRA achieves slightly lower performance on the target datasets (TSI and GTS) than \ours, but suffers a larger margin of forgetting on other datasets. SPU maintains the performance on ImgNet and DA·E similar to \ours but underperforms on the target dataset especially with TSI. LN accuracy on the target dataset is the lowest, however the accuracy on other dataset are also severely affected. 
\ours achieves the best performance on all datasets.

% After updating the CLIP models, we plug back the vision encoders into the respected VLMs and evaluate the VQA performance. Note that parameters of the VLM other than the vision encoder remain untouched. 
% Tables \ref{table:vlm_vqa_acc_tsi_offline} and \ref{table:vlm_vqa_acc_gtsrb_offline} report the VQA accuracies on TSI-VQA and GSTB-VQA respectively. It is clear that the VLM performance on the target dataset is significantly enhanced after adapting to TSI and GTSRB.  However, the effect on other datasets is less clear here and all methods perform similarly, probably as the LLM can still account for the small changes in the vision encoder. 
\myparagraph{Continual few-shot updates.}
Having evaluated the performance of \ours and other methods on the offline updates setup, we explore a more challenging setting: multiple consecutive sessions of updates with small numbers of examples in each.  The goal is to examine the robustness of \ours compared to other methods when continual few-shot updates are performed.  Table~\ref{table:clip_acc_tsi_fs20_cl} reports the accuracy on each test set at the end of each sequence for TSI and GTS with 20 shots in each session. Please refer to the Appendix for results with 50 shots.
Due the challenging nature of this setting, we see larger gaps among the compared methods. F-FT fails to achieve the best accuracy on the target dataset. It suffers from catastrophic forgetting, resulting in performance deterioration on the ImgNet and DALL·E datasets when training sequentially on TSI.   Interestingly, LoRA here shows a much weaker performance than the offline setting with accuracy on the target dataset close or worse than  finetuning and an even larger rates of forgetting on ImgNet and DA·E, especially after training on TSI. LN here remains behind \ours but achieves better performance than LoRA.  Low rank updates do not necessarily result in a reduced change in parameters and can significantly change the behaviour of the model on other datasets.  While SPU remains robust and achieves good performance on both target and other datasets, our method \ours improves on the target dataset with a considerable margin and has the least negligible deterioration (<2\%) on ImgNet and DA·E. Results are consistent for both  models and for both sequences (TSI and GTS). Note that SPU performance drops by more than 10\% on DA·E for the EVA-CLIP backbone under TSI sequence. 

\myparagraph{VQA performance after continual fewshot updates of CLIP.}
We have so far only reported the accuracy of finetuned CLIP on the classifications task. Here we investigate how plugging this continually finetuned models in the respective VLMs affect the VQA tasks. 
VQA results are reported in Table~\ref{table:vlm_vqa_acc_tsi_fs20_cl} for TSI and GTS under the 20-shot continual learning setting. 
For TSI, all methods were able to improve the performance over Zr-Shot, with \ours performing best, and with a margin of $2.6\%$ over LLaVa.  Our method also shows the least degradation on other VQA datasets, followed by SPU. Our method, \ours, even improves the performance on other VQA datasets for MiniGPTV2 with a small margin.
With respect to GTS, our method was the only method to improve the performance of VQA on GTS.  Other methods perform  worse than Zr-shot, except from SPU on LLaVA with similar performance to Zr-shot.  The performance on other datasets is less pronounced than the case of TSI. We attribute this to the out-of-distribution nature and the finegrained classes of GTS.  Further, the LLM seems to be able to cover for the changes in the CLIP backbones and suffers less disturbance on other datasets than when evaluating the classification performance of CLIP itself.
Overall, it can be stated that \ours is able to improve the performance on the VQA tasks for both  VLMs, even in a few-shot setting, and under different sequential sessions with no experience replay.
\begin{table}
\begin{center}
\begingroup
%\setlength{\tabcolsep}{2.1pt} % Default value: 6pt
%\renewcommand{\arraystretch}{.9}
%\begin{small}
%\begin{sc}
  \resizebox{1\textwidth}{!}
  {
\begin{tabular}{l c c c c c c c c c c c}
\toprule
&\multicolumn{5}{c}{\textbf{TSI}}&\multicolumn{5}{c}{\textbf{GTS}}\\
\midrule
 Method/Datast& {VSR}  & {VizWiz}  & {HM} & {DA-E} & \textbf{TSI} &|&{VSR}  & {VizWiz}  & {HM} & {DA-E} & \textbf{GTS} \\
\midrule
&\multicolumn{10}{c}{\textbf{MiniGPTv2}} \\
\midrule
Zr-Shot & $64.2$  & $56.6$ & $58.5$ &  $83.6$ & $62.9$&|& $64.2$  & $56.6$ & $58.5$ &  $83.6$ & $44.9$\\
F-FT & $62.4$ & $54.7$ & $57.1$ & $78.6$ & \underline{$71.7$}&|&$62.7$ & $56.4$ & $57.2$ & ${84.4}$ & $38.9$ \\
 LN &  $59.6$ & $52.0$ & $59.6$ & $76.3$ & $67.3$&|&$63.1$ & $56.8$ & $59.0$ & $83.8$ &\underline{$44.7$} \\
LoRA & $61.4$ & $52.7$ & $56.5$ & $71.8$ & $69.4$&|& $61.9$ & $55.8$ & $59.4$ & $82.7$ & $35.2$  \\
SPU & $62.7$ & $55.5$ & $59.0$ & $84.1$ & $71.1$& |&$63.3$ & $56.8$ & $58.6$ & $84.1$ & $43.4$ \\
LoRSU & $64.2$ & $59.6$ & $58.8$ & $84.0$ & $\mathbf{71.9}$ &|& $63.3$ & $56.7$ & $58.1$ & $84.1$ & $\mathbf{46.1}$\\
\midrule
&\multicolumn{10}{c}{\textbf{LLaVA}} \\
\midrule

Zr-Shot & $51.5$  & $63.5$ & $61.2$ &  $91.1$ & $53.1$& |&$51.5$  & $63.5$ & $61.2$ &  $91.1$ &{$75.6$}\\
F-FT & $51.6$ & $64.1$ & $61.3$ & $88.6$ & $65.9$& |&$62.7$ & $56.4$ & $57.2$ & $84.4$ & $38.9$  \\
 LN & $51.6$ & $57.3$ & $54.9$ & $70.3$ & $56.6$&|& $51.6$ & $63.0$ & $62.6$ & $89.7$ & $72.8$ \\
LoRA & $51.6$ & $61.1$ & $61.2$ & $77.4$ & $59.1$ &|&$51.6$ & $60.9$ & $61.2$ & $85.2$ & $46.7$\\
SPU & $51.7$ & $63.9$ & $60.5$ & $87.9$ & \underline{$66.5$} &|&$51.5$ & $63.3$ & $61.2$ & $91.5$ & \underline{$75.7$}   \\
LoRSU & $51.7$ & $63.1$ & $62.1$ & $90.0$ & $\mathbf{69.1}$ & |&$51.5$ & $63.2$ & $61.2$ & $91.2$ & $\mathbf{76.9}$\\
\bottomrule
\end{tabular}}
%\end{sc}
%\end{small}
\endgroup
\end{center}
\caption{ \footnotesize Accuracy (\%) scores for MiniGPTv2/LLaVA with the pretrained or fine-tuned CLIP EVA-ViT-G-14/OpenAI-CLIP-L-14 model. The vision encoder is \textit{separately} fine-tuned on each session (5 in total) of \textbf{TSI} (left) and \textbf{GTS} (right)  using \textbf{20 shots} per session, best method on the target dataset is highlighted in bold,  second best method is underscored. Our  \ours achieves the best accuracy on the target dataset.} 
\label{table:vlm_vqa_acc_tsi_fs20_cl}
\vspace*{-0.5cm}
\end{table}
% accuracy (\%)  scores for CLIP EVA-ViT-G-14 and OpenAI-ViT-L-14  models for various fine-tuning methods. The model is fine-tuned on each session (5 in total) of \textbf{TSI} (left) and \textbf{GTS} (right)  using \textbf{20 shots} per session, best method on the target dataset is highlighted in bold,  second best method is underscored. Our  \ours achieves the best performance on the target dataset with minimal deterioration on other datasets.

% \hdashline
\section{Limitations}
We considered the question of whether vision encoders can be updated separately to improve VLMs performance. Our  conclusions are made primarily based on the TSI and GTS datasets that were chosen to avoid using other publicly available datasets that have been used during CLIP pre-training. Further, we do not consider longer sequences due to compute resources, however interesting observations might follow. Our method is only applied to the ViT model, while in principle it can be integrated to update the LLM. We leave this for future work.  Studying the effect of separate updates in the context of the composite nature of VLMs is challenging and we have only scratched the surface.
\section{Conclusion}\label{sec:conclusion}
In this work, we investigated the limitations of CLIP on out-of-distribution benchmarks and found despite robustness on public and AI generated data  it  can fall short in certain scenarios with challenging visual conditions. These limitations are then inherited by the VLMs that utilize CLIP’s embeddings. To address this, we propose a novel approach: updating the vision encoder separately, specifically on data where CLIP fails. Remarkably, this strategy significantly corrects VLM mistakes on previously unseen images from the same data.  We further introduce a parameter-efficient tuning method that not only targets efficiency but also ensures the preservation of the model’s generic knowledge. In our experiments, our method, \ours, is the only method to systematically improve the VLM performance on VQA tasks even in the challenging, but realistic, continual fewshot setting.  Our approach hence strikes a strong balance between efficiency, effectiveness and robustness, achieving new state-of-the-art results.

\bibliography{main}
\bibliographystyle{plain}

%%%%%%%%%%%%%%%%%%%%%%%%%%%%%%%%%%%%%%%%%%%%%%%%%%%%%%%%%%%%%%%%%%%%%%%%%%%%%%%
%%%%%%%%%%%%%%%%%%%%%%%%%%%%%%%%%%%%%%%%%%%%%%%%%%%%%%%%%%%%%%%%%%%%%%%%%%%%%%%
% APPENDIX
%%%%%%%%%%%%%%%%%%%%%%%%%%%%%%%%%%%%%%%%%%%%%%%%%%%%%%%%%%%%%%%%%%%%%%%%%%%%%%%
%%%%%%%%%%%%%%%%%%%%%%%%%%%%%%%%%%%%%%%%%%%%%%%%%%%%%%%%%%%%%%%%%%%%%%%%%%%%%%%
\newpage
\appendix

\section{Appendix}

\subsection{Additional  Results}
Due to space limit, we report here the results of other experimental setting. First, we report VQA after offline updates of CLIP. In Table~\ref{table:vlm_vqa_acc_tsi_offline} and Table~\ref{table:vlm_vqa_acc_gtsrb_offline} we report the VQA for both considered VLMs after finetuning CLIP models on TSI and GTS datasets respectively. Our method achieves the best performance on the target dataset compared to other parameter efficient tuning methods while maintaining a strong performance on other VQA datasets.
Next we considered a slightly more relaxed continual few-shot setting, by allowing 50 shots per session.
Tables~\ref{table:clip_acc_tsi_fs50_cl} and ~\ref{table:clip_acc_gstrb_fs50_cl} report the classification accuracy of the two considered CLIP models after fewshot continual learning on TSI and GTS datasets. Similar conclusions can be made here. Our methods continue to perform the best on the target dataset while maintaining the performance on other datasets. Our method specifically set a large margin of 16\% on DA·E compared to other methods when the target dataset is TSI. This indicates that our method updates are generalizable to other images of the same actions differently from other studied methods.
\subsection{Prompts used to generate images from DALL·E 2}
We generated images from DALL·E 2 using OpenAI python package and we used the prompt
 ``\textit{A person} $\{a\}$'' where $a \in $ \{ \textit{using a white coffee machine, 
                 eating, 
                 cutting bread, 
                 stirring the pot, 
                 holding a glass, 
                 watching TV, 
                 holding a bottle, 
                 walking, 
                 making tea, 
                 cutting food, 
                 holding a cup, 
                 using a laptop, 
                 lying down, 
                 holding a can, 
                 person holding a black kettle, 
                 reading a book, 
                 cleaning up, 
                 sitting down, 
                 using a tablet, 
                 boiling water in a black kettle, 
                 using a cordless phone, 
                 washing dishes}\}.

\newpage
% Optionally include supplemental material (complete proofs, additional experiments and plots) in appendix.
% All such materials \textbf{SHOULD be included in the main submission.}
%\rahaf{ Ares we need to describe the hyper parameters tunning and the GPU used for training.}
\section{Implementation Details}\label{sec:implementation}
\begin{itemize}
    \item We use a single A100 GPU for the experiments.
    \item We use Adam as an optimizer with and learning rate scheduler of Cosine Annealing with Warmup for all methods.
    \item We use batch size 8 for the few shot experiments and bsize 64 for the offline ones.
    \item For few-shot experiments we use 50 epochs for the TSI dataset and 10 epochs for GTS. We use 20 epochs and 10 epochs for the TSI dataset and GTS, respectively for the offline setting.
    \item For Lora we use rank=32 for all experiments.
    \item For SPU we use sparsity=10\% for all experiments.
    \item For \ours we use sparsity=5\%, rank=32, and we pick the top-2 attention heads for all experiments.
    \item For \ours and SPU, to pick the top-k parameters from the first MLP layer we use either 800 for the offline setting or all the available data points in the dataset of the current task for CL-few shot setting.
    \item For all VQA datasets we measure performance based on accuracy of the predicted answers of the VLM.
    \item We converted TSI-VQA, GTS-VQA, and DA·E-VQA as a multiple choice VQA problem, each question has 5 options and the VLM is asked to choose the right one. For the other other datasets we follow \cite{chen2023minigpt} for the evaluation protocol. 
\end{itemize}

\begin{table}
\begin{center}
\begingroup
%\setlength{\tabcolsep}{2.1pt} % Default value: 6pt
%\renewcommand{\arraystretch}{.9}
%\begin{small}
%\begin{sc}
\begin{tabular}{l c }
\toprule
\textbf{Original Class name/Action} & \textbf{Generated Caption}  \\
\midrule
Cook.Cleandishes & washing dishes \\
Cook.Cleanup & cleaning up \\
Cook.Cut & cutting food \\
Cook.Stir & stirring the pot \\
Cook.Usestove & \xmark \\
Cook.Cutbread & cutting bread \\
Drink.Frombottle & holding a bottle \\
Drink.Fromcan & holding a can \\
Drink.Fromcup & holding a cup \\
Drink.Fromglass & holding a glass \\
Eat.Attable & eating \\
Eat.Snack & \xmark \\
Enter & walking \\
Getup & \xmark \\
Laydown & lying down \\
Leave & walking \\
Makecoffee.Pourgrains & using a white coffee machine \\
Makecoffee.Pourwater & using a white coffee machine \\
Maketea.Boilwater & boiling water in a black kettle \\
Maketea.Boilwater & making tea \\
Maketea.Insertteabag & making tea \\
Pour.Frombottle & holding a bottle \\
Pour.Fromcan & holding a can \\
Pour.Fromkettle & holding a black kettle \\
Readbook & reading a book \\
Sitdown & sitting down \\
Takepills & \xmark \\
Uselaptop & using a laptop \\
Usetablet & using a tablet \\
Usetelephone & using a cordless phone \\
Walk & walking \\
WatchTV & watching TV \\
\bottomrule
\end{tabular}
%\end{sc}
%\end{small}
\endgroup
\end{center}
\caption{The original action names of the Toyota Smarthome dataset and their corresponding captions used to create the Toyota Smarthome Images (TSI) dataset. We use~\xmark~to denore the actions that are ambiguous and were not used to build the TSI dataset. The final prompt is created as ``\textit{The person in this image is \{caption\}}''.}
\label{table:tsi_class_names}
\end{table}
 \begin{table}
\begin{center}
\begingroup
%\setlength{\tabcolsep}{2.1pt} % Default value: 6pt
%\renewcommand{\arraystretch}{.9}
%\begin{small}
%\begin{sc}
\begin{tabular}{l c c c c c c}
\toprule
 & & \multicolumn{5}{c}{\textbf{VQA Datasets (Acc \%)}}  \\
 \cmidrule(lr){3-7}
\textbf{Setting} & \textbf{PEFT Method} & \textbf{VSR}  & \textbf{VizWiz}  & \textbf{HM} & \textbf{DA·E} & \textbf{TSI} \\
\midrule
\multirow{2}{*}{\textbf{CL-20 shots}} & \textbf{LoRA (LLM)} & $63.3$ & $54.3$  & $58.4$ & $88.8$ &  $71.6$  \\
 & \textbf{LoRSU (EVA-ViT-G)} & $64.2$ & $59.6$ & $58.8$ & $84.0$ & $71.9$  \\
 \midrule
\midrule
\multirow{2}{*}{\textbf{CL-50 shots}} & \textbf{LoRA (LLM)} & $64.2$ & $54.9$  & $58.3$ &  $88.0$ & $73.2$ \\
 & \textbf{LoRSU (EVA-ViT-G)} & $63.6$ & $56.3$ & $57.5$ & $83.6$ & $73.4$ \\
  \midrule
\midrule
\multirow{2}{*}{\textbf{Offline}} & \textbf{LoRA (LLM)} & $63.6$ & $55.3$  & $57.6$ & $86.5$ &   $82.3$ \\
 & \textbf{LoRSU (EVA-ViT-G)} & $62.4$ & $56.4$ & $58.7$ & $86.2$ & $83.1$ \\
\bottomrule
\end{tabular}
%\end{sc}
%\end{small}
\endgroup
\end{center}
\caption{Accuracy (\%) scores for MiniGPTv2. We fine-tune the LLM using LoRA (LoRA (LLM)) with $r=64$ on the \textbf{TSI} dataset under different settings (the visual encoder remains frozen) and we compare its performance to our method LoRSU that fine-tunes the visual encoder.} 
\label{table:fine_tune_llm_tsi}
\end{table}

 \begin{table}
\begin{center}
\begingroup
%\setlength{\tabcolsep}{2.1pt} % Default value: 6pt
%\renewcommand{\arraystretch}{.9}
%\begin{small}
%\begin{sc}
\begin{tabular}{l c c c c c c}
\toprule
 & & \multicolumn{5}{c}{\textbf{VQA Datasets (Acc \%)}}  \\
 \cmidrule(lr){3-7}
\textbf{Setting} & \textbf{PEFT Method} & \textbf{VSR}  & \textbf{VizWiz}  & \textbf{HM} & \textbf{DA·E} & \textbf{GTS} \\
\midrule
\multirow{2}{*}{\textbf{CL-20 shots}} & \textbf{LoRA (LLM)} & $61.9$ & $56.0$ & $57.9$ & $84.6$ & $33.6$  \\
 & \textbf{LoRSU (EVA-ViT-G)} & $63.3$ & $56.7$ & $58.1$ & $84.1$ & $46.1$  \\
 \midrule
\midrule
\multirow{2}{*}{\textbf{CL-50 shots}} & \textbf{LoRA (LLM)} & $61.9$ & $56.3$  & $57.8$ &  $84.7$ & $33.8$ \\
 & \textbf{LoRSU (EVA-ViT-G)} & $63.3$ & $56.7$ & $57.9$ & $84.1$ & $47.6$ \\
  \midrule
\midrule
\multirow{2}{*}{\textbf{Offline}} & \textbf{LoRA (LLM)} & $61.8$ & $55.9$  & $55.2$ & $84.7$ &  $86.2$ \\
 & \textbf{LoRSU (EVA-ViT-G)} & $62.9$ & $57.0$ & $58.8$ & $83.7$ & $53.5$ \\
\bottomrule
\end{tabular}
%\end{sc}
%\end{small}
\endgroup
\end{center}
\caption{Accuracy (\%) scores for MiniGPTv2. We fine-tune the LLM using LoRA (LoRA (LLM)) with $r=64$ on the \textbf{GTS} dataset under different settings (the visual encoder remains frozen) and we compare its performance to our method LoRSU that fine-tunes the visual encoder.} 
\label{table:fine_tune_llm_gts}
\end{table}

\begin{table}
\begin{center}
\begingroup
%\setlength{\tabcolsep}{2.1pt} % Default value: 6pt
%\renewcommand{\arraystretch}{.9}
%\begin{small}
%\begin{sc}
\begin{tabular}{l c c}
\toprule
\textbf{Visual Encoder (Total \#Params)} & \textbf{Method} & \textbf{Trainable \#Params} \\
\midrule
\multirow{5}{*}{\textbf{EVA-CLIP-G-14 (1012.6M)}} & LN  & $0.2M (0.02\%)$ \\
& F-FT  & $1012.6M (100\%)$ \\
& LoRA   & $30.2M (2.98\%)$  \\
& SPU   &  $34.6M (3.42\%)$  \\
& LoRSU-Ours  & $24.6M (2.43\%)$  \\
\midrule
\multirow{5}{*}{\textbf{OpenAI-CLIP-L-14 (304.3M)}}  & LN  & $0.1M (0.03\%)$  \\
& F-FT   & $304.3M (100\%)$ \\
& LoRA  & $12.6M (4.15\%)$  \\
& SPU   & $10.1M (3.31\%)$  \\
& LoRSU-Ours  & $8.2M (2.71\%)$  \\
\midrule
\bottomrule
\end{tabular}
%\end{sc}
%\end{small}
\endgroup
\end{center}
\caption{Parameter efficiency for each method considered in our experiments.} 
\label{table:number_parameters_method}
\end{table}
\begin{table}
\begin{center}
\begingroup
%\setlength{\tabcolsep}{2.1pt} % Default value: 6pt
%\renewcommand{\arraystretch}{.9}
%\begin{small}
%\begin{sc}
\begin{tabular}{l c c c}
\toprule
\textbf{Method} & \textbf{Minutes/epoch} & \textbf{TFlops (Forward)} & \textbf{Trainable Params (M)}\\
\midrule
LoRA-LLM & $49.6$  & $6.2$ & $33.6$ \\
LoRSU (EVA-ViT-G) & $9.8$ & $0.53$ & $24.6$ \\
\bottomrule
\end{tabular}
%\end{sc}
%\end{small}
\endgroup
\end{center}
\caption{TFlops and time comparison between the fine-tuned LLM (LoRA) and the fine-tuned visual encoder (LoRSU) for the \textbf{offline} setting on \textbf{GTS} dataset. We report results based on a single NVidia A100 GPU.} 
\label{table:gflops_comparison}
\end{table}

\begin{table}
\begin{center}
\begingroup
%\setlength{\tabcolsep}{2.1pt} % Default value: 6pt
%\renewcommand{\arraystretch}{.9}
%\begin{small}
%\begin{sc}
\begin{tabular}{l c c c c c c}
\toprule
   &  \multicolumn{3}{c}{\textbf{EVA-Clip-G-14}} & \multicolumn{3}{c}{\textbf{ OpenAI-Clip-L-14}}  \\
  \cmidrule(lr){2-4} \cmidrule(lr){5-7} \textbf{Dataset} & \textbf{ImgNet} & \textbf{DA·E} & \textbf{TSI}  & \textbf{ImgNet} & \textbf{DA·E} & \textbf{TSI} \\
\midrule
\textbf{Zr-Shot} & $78.5$ & $97.0$ & $11.4$ & $76.6$ & $90.9$ & $13.2$ \\
\midrule
\textbf{LN} & $33.2$ & $67.7$ & $39.7$ & $72.6$ & $88.6$ & $42.5$ \\
\midrule
\textbf{F-FT} & $47.9$ & $75.33$ & $38.6$ & $73.0$ & $88.5$ & $36.9$ \\
\midrule
\textbf{LoRA} & $32.8$ & $58.3$ & $36.4$ & $67.4$ & $73.9$ & $29.8$ \\
\midrule
\textbf{SPU} & $70.6$ & $77.9$ & $42.6$ & $76.6$ & $90.5$ & $56.4$ \\
\midrule
\textbf{LoRSU}-Ours & $77.8$ & $94.2$ & $48.6$ & $76.6$ & $90.6$ & $58.1$ \\
\bottomrule
\end{tabular}
%\end{sc}
%\end{small}
\endgroup
\end{center}
\caption{\textbf{Last session} accuracy for CLIP EVA-ViT-g-14 and OpenAI-ViT-L-14-336 models for various fine-tuning methods. The model is fine-tuned on each session (5 in total) of \textbf{TSI} dataset using \textbf{50 shots} per session.} 
\label{table:clip_acc_tsi_fs50_cl}
\end{table}
\begin{table}
\begin{center}
\begingroup
%\setlength{\tabcolsep}{2.1pt} % Default value: 6pt
%\renewcommand{\arraystretch}{.9}
%\begin{small}
%\begin{sc}
\begin{tabular}{l c c c c c c}
\toprule
\textbf{VLM} & \textbf{Method/Dataset} & \textbf{VSR}  & \textbf{VizWiz}  & \textbf{HM} & \textbf{DA-E} & \textbf{TSI} \\
\midrule
\multirow{6}{*}{\textbf{MiniGPTv2}} & Zr-Shot & $64.2$  & $56.6$ & $58.5$ &  $83.6$ & $62.9$\\
 & LN &  $57.8$ & $49.9$ & $56.8$ & $65.9$ & $68.3$ \\
& F-FT & $61.1$ & $53.4$ & $57.3$ & $74.6$ & $66.0$ \\
& LoRA & $52.5$ & $50.3$ & $59.9$ & $54.1$ & $70.1$ \\
& SPU & $64.0$ & $56.3$ & $59.3$ & $79.7$ & $71.8$ \\
& LoRSU-Ours & $63.6$ & $56.3$ & $57.5$ & $83.6$ & $73.4$ \\
\midrule
\multirow{6}{*}{\textbf{LLaVA}} & Zr-Shot & $51.5$  & $63.5$ & $61.2$ &  $91.1$ & $53.1$\\
 & LN &  $52.1$ & $53.5$ & $53.6$ & $55.6$ & $64.7$ \\
& F-FT & $51.6$ & $62.7$ & $61.9$ & $78.3$ & $69.0$ \\
& LoRA & $52.0$ & $53.7$ & $53.1$ & $39.1$ & $65.0$ \\
& SPU & $51.6$ & $64.1$ & $61.1$ & $87.0$ & $70.3$ \\
& LoRSU-Ours & $51.6$ & $63.2$ & $60.8$ & $88.9$ & $71.6$ \\
\midrule
\bottomrule
\end{tabular}
%\end{sc}
%\end{small}
\endgroup
\end{center}
\caption{Accuracy (\%) scores for MiniGPTv2/LLaVA with the pretrained or fine-tuned CLIP EVA-ViT-g-14/OpenAI-CLIP-L-14-336px model. All fine-tuning methods use \textbf{TSI} data to fine-tune the visual encoder for \textbf{50 shots} in 5 sessions. }%The methods whose results are not available due to limited compute resources are denoted by $-$.}
\label{table:vlm_vqa_acc_tsi_fs50_cl}
\end{table}
\begin{table}
\begin{center}
\begingroup
%\setlength{\tabcolsep}{2.1pt} % Default value: 6pt
%\renewcommand{\arraystretch}{.9}
%\begin{small}
%\begin{sc}
\begin{tabular}{l c c c c c c}
\toprule
\textbf{VLM} & \textbf{Method/Dataset} & \textbf{VSR}  & \textbf{VizWiz}  & \textbf{HM} & \textbf{DA·E} & \textbf{TSI} \\
\midrule
\multirow{6}{*}{\textbf{MiniGPTv2}} & Z-Shot & $64.2$  & $56.6$ & $58.5$ &  $83.6$ & $62.9$\\
 & LN &  $62.2$ & $55.7$ & $59.0$ & $81.2$ & $70.2$ \\
& F-FT & $61.9$ & $55.3$ & $59.5$ & $87.1$ & $86.0$ \\
& LoRA & $61.5$ & $55.6$ & $58.4$ & $85.8$ & $80.2$ \\
& SPU & $63.1$ & $56.7$ & $57.8$ & $85.3$ & $73.5$ \\
& LoRSU-Ours & $62.4$ & $56.4$ & $58.7$ & $86.2$ & $83.1$ \\
\midrule
\multirow{6}{*}{\textbf{LLaVA}} & Zr-Shot & $51.5$  & $63.5$ & $61.2$ &  $91.1$ & $53.1$\\
 & LN &  $51.7$ & $62.9$ & $61.1$ & $87.9$ & $74.3$ \\
& F-FT & $51.8$ & $63.9$ & $61.8$ & $89.9$ & $81.0$ \\
& LoRA & $51.6$ & $63.8$ & $61.0$ & $89.7$ & $80.2$ \\
& SPU & $51.5$ & $63.6$ & $62.0$ & $91.1$ & $71.7$ \\
& LoRSU-Ours & $51.4$ & $63.7$ & $61.9$ & $91.1$ & $80.6$ \\
\midrule
\bottomrule
\end{tabular}
%\end{sc}
%\end{small}
\endgroup
\end{center}
\caption{Accuracy (\%) scores for MiniGPTv2/LLaVA with the pretrained or fine-tuned CLIP EVA-ViT-g-14/OpenAI-CLIP-L-14-336px model. All fine-tuning methods use the whole training split (\textbf{offline}) of \textbf{TSI} data to fine-tune the visual encoder.}
\label{table:vlm_vqa_acc_tsi_offline}
\end{table}

\begin{table}
\begin{center}
\begingroup
\begin{tabular}{l c c c c c c}
\toprule
   &  \multicolumn{3}{c}{\textbf{EVA-Clip-G-14}} & \multicolumn{3}{c}{\textbf{ OpenAI-Clip-L-14}}  \\
  \cmidrule(lr){2-4} \cmidrule(lr){5-7} \textbf{Dataset} & \textbf{ImgNet} & \textbf{DA-E} & \textbf{GTS}  & \textbf{ImgNet} & \textbf{DA-E} & \textbf{GTS} \\
\midrule
\textbf{Zr-Shot} & $78.5$ & $97.0$ & $49.1$ & $76.6$ & $90.9$ & $52.4$ \\
\midrule
\textbf{LN} & $76.8$ & $97.3$ & $49.0$ & $72.6$ & $88.6$ & $42.5$ \\
\midrule
\textbf{F-FT} & $77.0$ & $96.2$ & $36.9$ & $73.0$ & $88.5$ & $36.9$ \\
\midrule
\textbf{LoRA} & $74.7$ & $93.2$ & $35.7$ & $64.4$ & $73.9$ & $29.8$ \\
\midrule
\textbf{SPU} & $78.5$ & $96.8$ & $54.9$ & $76.6$ & $90.5$ & $56.7$ \\
\midrule
\textbf{LoRSU}-Ours & $78.5$ & $96.8$ & $57.5$ & $76.6$ & $90.5$ & $58.2$ \\
\bottomrule
\end{tabular}
%\end{sc}
%\end{small}
\endgroup
\end{center}
\caption{\textbf{Last session} accuracy for CLIP EVA-ViT-g-14 and OpenAI-ViT-L-14-336 models for various fine-tuning methods. The model is fine-tuned on each session (5 in total) of \textbf{GTS} dataset using \textbf{50 shots} per session.} 
\label{table:clip_acc_gstrb_fs50_cl}
\end{table}
\begin{table}
\begin{center}
\begingroup
%\setlength{\tabcolsep}{2.1pt} % Default value: 6pt
%\renewcommand{\arraystretch}{.9}
%\begin{small}
%\begin{sc}
\begin{tabular}{l c c c c c c}
\toprule
\textbf{VLM} & \textbf{Method/Dataset} & \textbf{VSR}  & \textbf{VizWiz}  & \textbf{HM} & \textbf{DA-E} & \textbf{GTS} \\
\midrule
\multirow{6}{*}{\textbf{MiniGPTv2}} & Zr-Shot & $64.2$  & $56.6$ & $58.5$ &  $83.6$ & $44.9$\\
 & LN & $62.3$ & $56.2$ & $58.9$ & $82.9$ & $44.1$ \\
& F-FT & $62.5$ & $56.5$ & $58.6$ & $83.8$ & $34.5$ \\
& LoRA & $62.4$ & $54.3$ & $61.0$ & $82.1$ & $33.0$ \\
& SPU & $62.9$ & $56.5$ & $58.5$ & $83.8$ & $44.5$ \\
& LoRSU-Ours & $63.3$ & $56.7$ & $57.9$ & $84.1$ & $47.6$ \\
\midrule
\multirow{6}{*}{\textbf{LLaVA}} & Zr-Shot & $51.5$  & $63.5$ & $61.2$ &  $91.1$ & $75.6$\\
 & LN & $51.6$ & $63.4$ & $62.6$ & $91.5$ & $67.2$  \\
& F-FT & $51.6$ & $60.4$ & $65.3$ & $87.9$ & $54.3$ \\
& LoRA & $51.6$ & $58.1$ & $61.4$ & $80.8$ & $45.0$ \\
& SPU & $51.5$ & $63.1$ & $61.5$ & $91.7$ & $76.6$ \\
& LoRSU-Ours & $51.5$ & $63.2$ & $61.4$ & $91.7$ & $77.4$ \\
\midrule
\bottomrule
\end{tabular}
%\end{sc}
%\end{small}
\endgroup
\end{center}
\caption{Accuracy (\%) scores for MiniGPTv2/LLaVA with the pretrained or fine-tuned CLIP EVA-ViT-g-14/OpenAI-CLIP-L-14-336px model. All fine-tuning methods use \textbf{GTS} data to fine-tune the visual encoder for \textbf{50 shots} in 5 sessions.} 
\label{table:vlm_vqa_acc_gtsrb_fs50_cl}
\end{table}
\begin{table}
\begin{center}
\begingroup
%\setlength{\tabcolsep}{2.1pt} % Default value: 6pt
%\renewcommand{\arraystretch}{.9}
%\begin{small}
%\begin{sc}
\begin{tabular}{l c c c c c c}
\toprule
\textbf{VLM} & \textbf{Method/Dataset} & \textbf{VSR}  & \textbf{VizWiz}  & \textbf{HM} & \textbf{DA·E} & \textbf{GTS} \\
\midrule
\multirow{6}{*}{\textbf{MiniGPTv2}} & Zr-Shot & $64.2$  & $56.6$ & $58.5$ &  $83.6$ & $44.9$\\
 & LN & $61.4$ & $55.1$ & $58.4$ & $80.8$ & $52.7$ \\
& F-FT & $61.7$ & $56.3$ & $57.2$ & $83.3$ & $55.2$ \\
& LoRA & $62.0$ & $56.5$ & $58.9$ & $83.9$ & $50.8$ \\
& SPU & $63.3$ & $57.1$ & $59.1$ & $83.8$ & $49.5$ \\
& LoRSU-Ours & $62.9$ & $57.0$ & $58.8$ & $83.7$ & $53.5$ \\
\midrule
\multirow{6}{*}{\textbf{LLaVA}} & Zr-Shot & $51.5$  & $63.5$ & $61.2$ &  $91.1$ & $75.6$\\
 & LN & $51.6$ & $60.6$ & $61.2$ & $83.3$ & $85.0$  \\
& F-FT & $51.6$ & $63.6$ & $62.1$ & $90.5$ & $91.8$ \\
& LoRA & $51.6$ & $64.1$ & $62.4$ & $88.5$ & $88.2$ \\
& SPU & $51.6$ & $63.5$ & $61.3$ & $91.2$ & $87.0$ \\
& LoRSU-Ours & $51.6$ & $64.1$ & $62.4$ & $91.2$ & $91.5$ \\
\midrule
\bottomrule
\end{tabular}
%\end{sc}
%\end{small}
\endgroup
\end{center}
\caption{Accuracy (\%) scores for MiniGPTv2/LLaVA with the pretrained or fine-tuned CLIP EVA-ViT-g-14/OpenAI-CLIP-L-14-336px model. All fine-tuning methods use the whole training split (\textbf{offline}) of \textbf{GTS} data to fine-tune the visual encoder. }%The methods that results are not available due to limited compute resources are denoted by $-$.} 
\label{table:vlm_vqa_acc_gtsrb_offline}
\end{table}

\section{Parameters efficiency}
Table~\ref{table:number_parameters_method} reports the number of parameters updated by each method and the percentage with respect to model size for both considered CLIP models.  LN uses the least amount of parameters, however it lacks behind in accuracy on all evaluated datasets. \ours operates on fewer parameters compared to LoRa and SPU and yet strikes a strong balance between target datasets and the maintenance of generic knowledge, achieving the best performance in both classification and VQA tasks. 
\section{TSI Datset construction}
 To extract  images from the videos of the Toyota Smart Home dataset~ (TSI), we discretized each video clip into 2 frames per second and then selected the frame in the middle of the total time duration of the video clip. In Table~\ref{table:tsi_class_names} we describe the actions that were selected and the corresponding prompt used for CLIP classification. We also note dropping few actions to avoid ambiguous classes. 
 \section{Evaluation of CLIP on XImageNet-12}\label{sec:ximagenet}
In the Section~\ref{sec:clip-robustness} we evaluated CLIP robustness on XImageNet-12 benchmark~\cite{li2023ximagenet}. Here we describe this experiment in more detail.
XImageNet-12 benchmark~\cite{li2023ximagenet} covers 12 common categories from ImageNet and simulating six diverse out of distribution effects, such as overexposure, blurring, and color changing.  
 Table~\ref{tab-ximagenet} reports the results of CLIP ViT-B-16 with different pretraining. Although only one domain with random backgrounds of other objects exhibits weak performance, this could be attributed to model confusion between the two objects in the foreground and background, rather than a weakness in understanding the image.

\begin{figure}
\center
  \begin{minipage}[b]{0.85\linewidth} 
   \resizebox{1\textwidth}{!}{
 \begin{tabular}{l c c c c c }
\toprule
 &  \multicolumn{5}{c}{\textbf{Test Dataset (Top-1 Acc \%)}}  \\
\cmidrule(lr){2-6}
 \textbf{Pretraining Dataset}   &  \textbf{Blur\_bg} & \textbf{Blur\_obj} & \textbf{Color} & \textbf{Rand\_bg} & \textbf{Seg\_img} \\
\midrule
ViT-B-16 ( ImageNet) &  $88.4$ & $ 90.8$ &$66.5$&  $17.2$ & $49.0$ \\
ViT-B-16 (XImageNet-12)
 & $71.51$ & $70.21$ &$74.14$ & $38.01$ & $78.7$ \\
\midrule
CLIP-ViT-B-16 (DATACOMP) & $98.9$ & $97.5$ &$98.6$ & $42.4$  & $95.4$ \\
\midrule
CLIP-ViT-L-14 (OpenAI) & $98.9$ & $98.2$ &$98.3$ & $52.5$  & $95.7$ \\
\bottomrule
\end{tabular}}
\captionof{table}{\label{tab-ximagenet}\footnotesize  Performance on XImageNet-12 benchmark with ViT-B and ViT-L considering  different pretraining settings. CLIP pretraining with DATACOMP is quite robust to various shifts. }%\ares{The results for ImageNet* and XImageNet-12 are based on a ViT-B-16 backbone}
\end{minipage}   
\end{figure}

\subsection{Examples of TSI and DALL·E (DA·E) datasets}
\begin{figure}%
    \centering
    \subfloat[\centering TSI]{{\includegraphics[width=.4\linewidth]{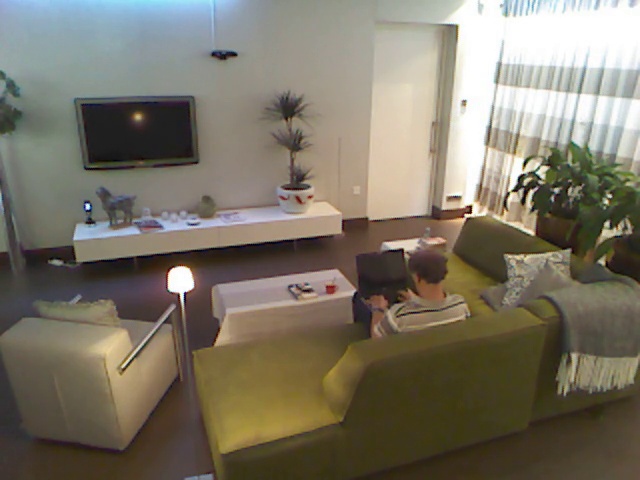}}}
    \qquad
    \subfloat[\centering DALL·E]{{\includegraphics[width=.4\linewidth]{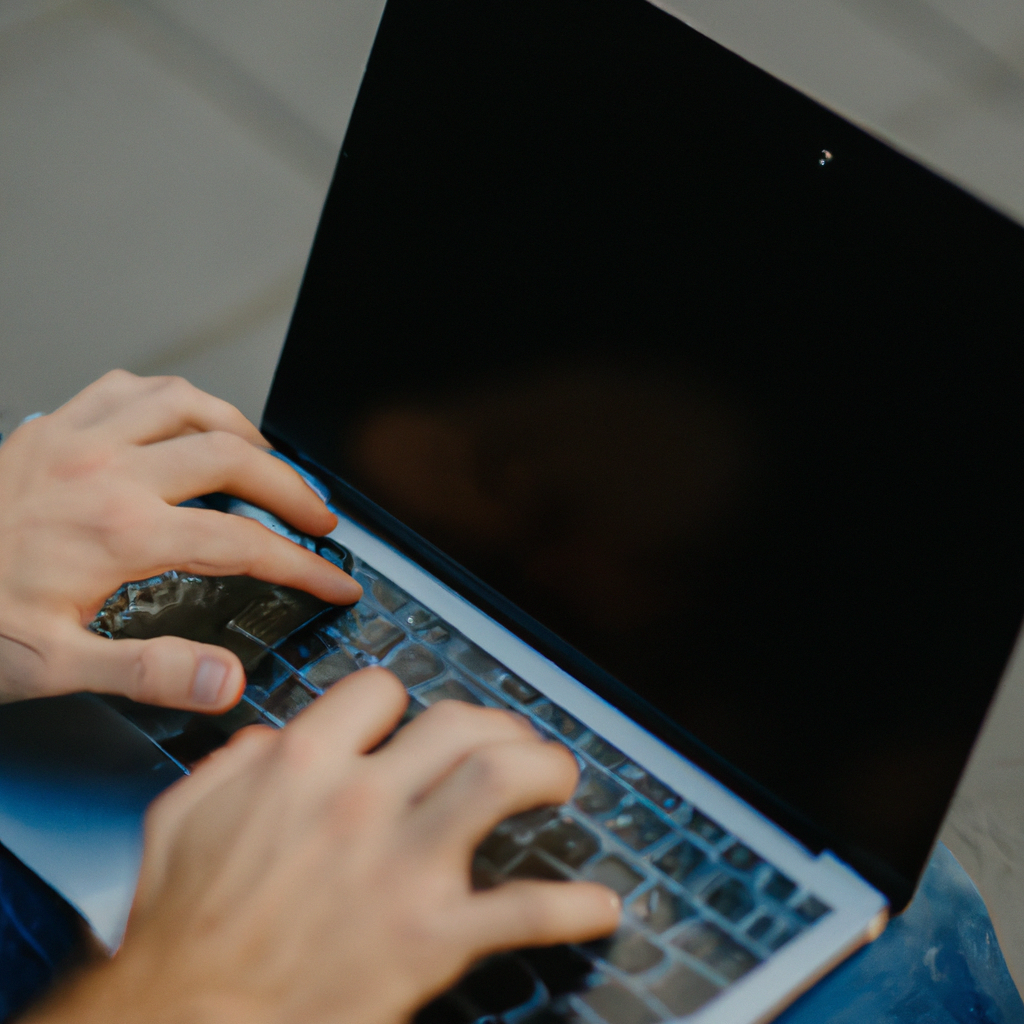}}}
    \caption{Use Laptop Example}%
    \label{fig:example1}%
\end{figure}

\begin{figure}%
    \centering
    \subfloat[\centering TSI]{{\includegraphics[width=.4\linewidth]{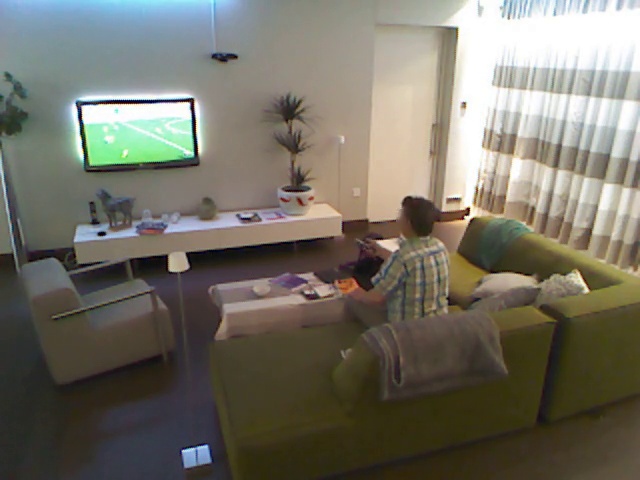}}}
    \qquad
    \subfloat[\centering DALL·E]{{\includegraphics[width=.4\linewidth]{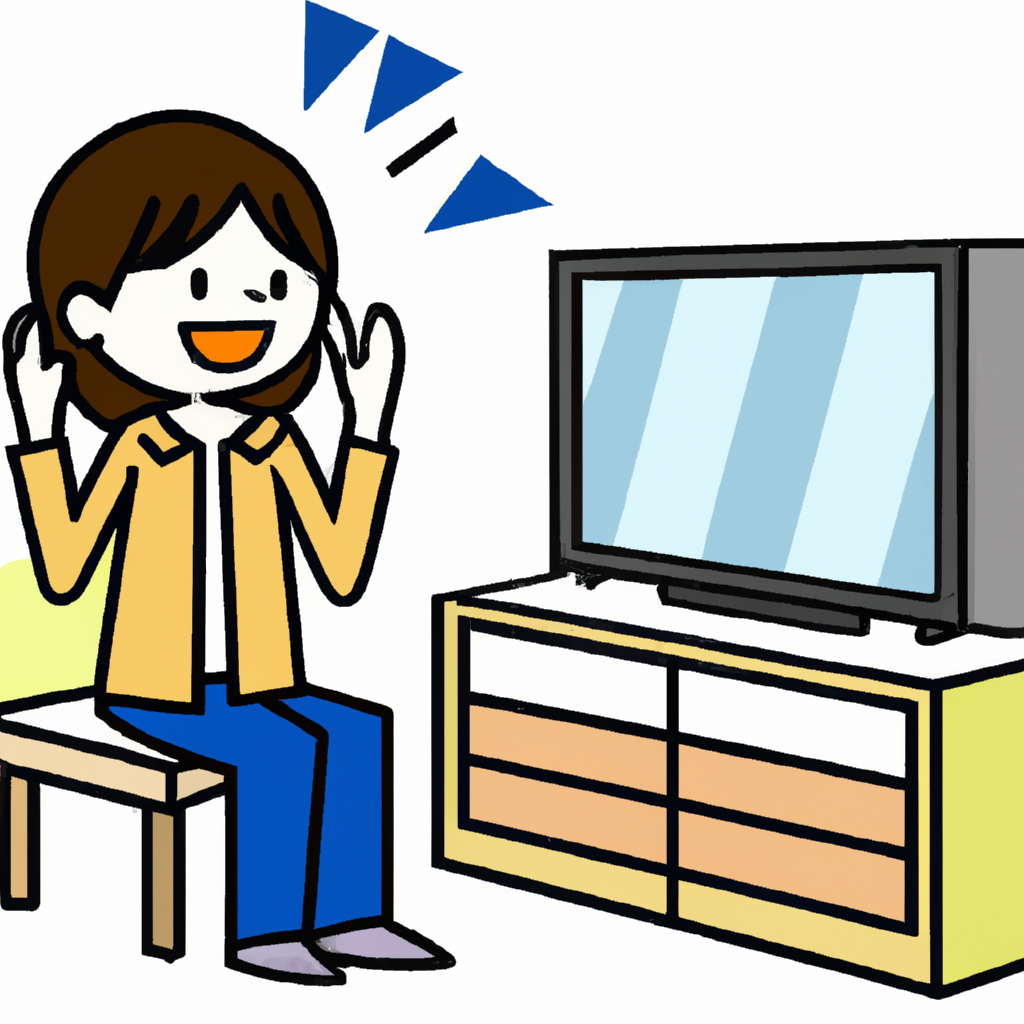}}}%
    \caption{Watching TV Example}%
    \label{fig:example2}%
\end{figure}

\begin{figure}%
    \centering
    \subfloat[\centering TSI]{{\includegraphics[width=.4\linewidth]{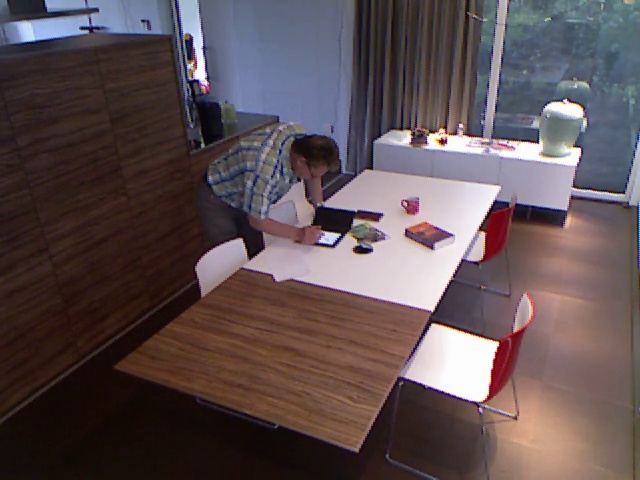}}}%
    \qquad
    \subfloat[\centering DALL·E]{{\includegraphics[width=.4\linewidth]{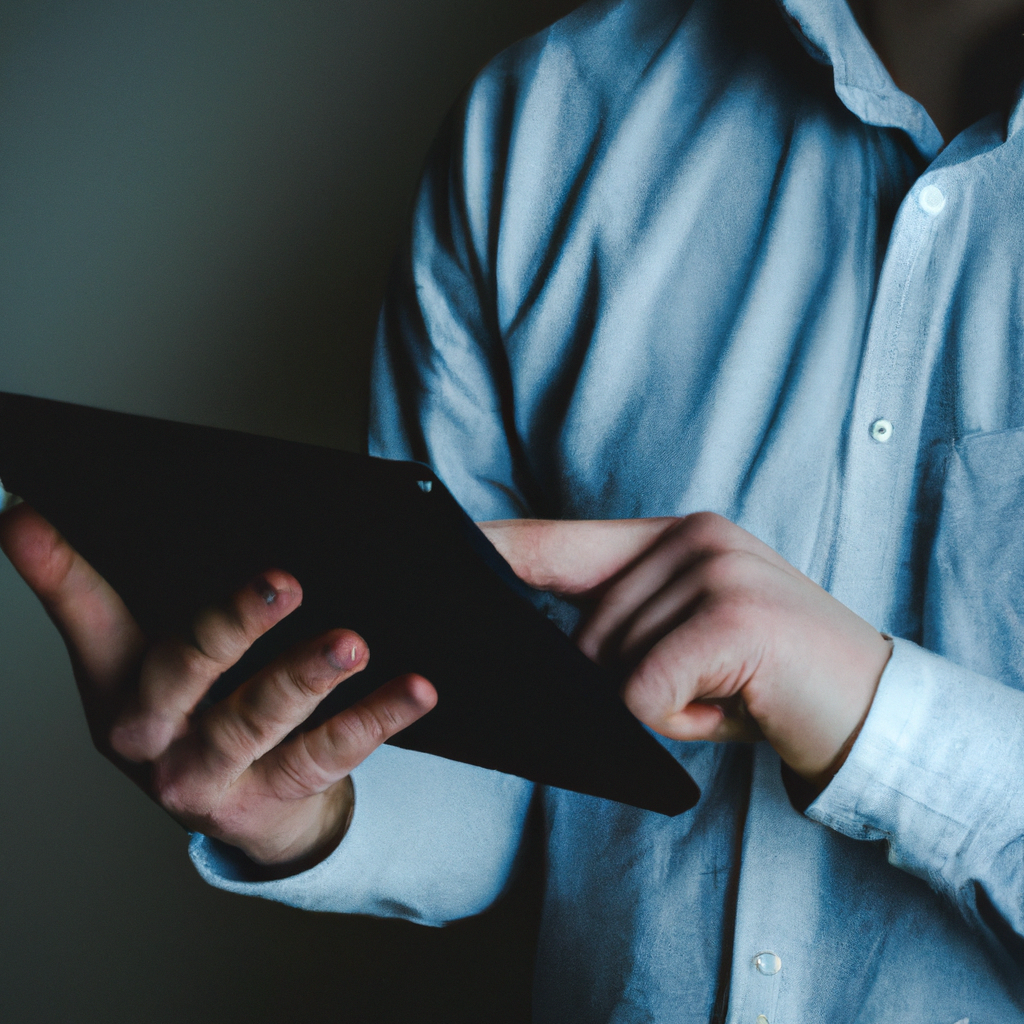}}}%
    \caption{Use Tablet Example}%
    \label{fig:example3}%
\end{figure}

\begin{figure}%
    \centering
    \subfloat[\centering TSI]{{\includegraphics[width=.4\linewidth]{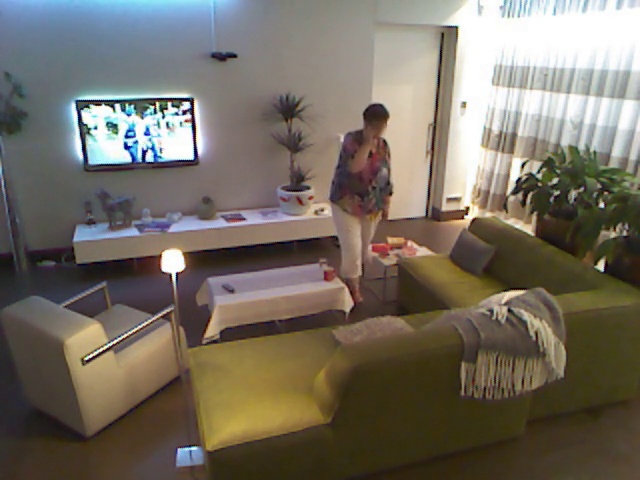}}}%
    \qquad
    \subfloat[\centering DALL·E]{{\includegraphics[width=.4\linewidth]{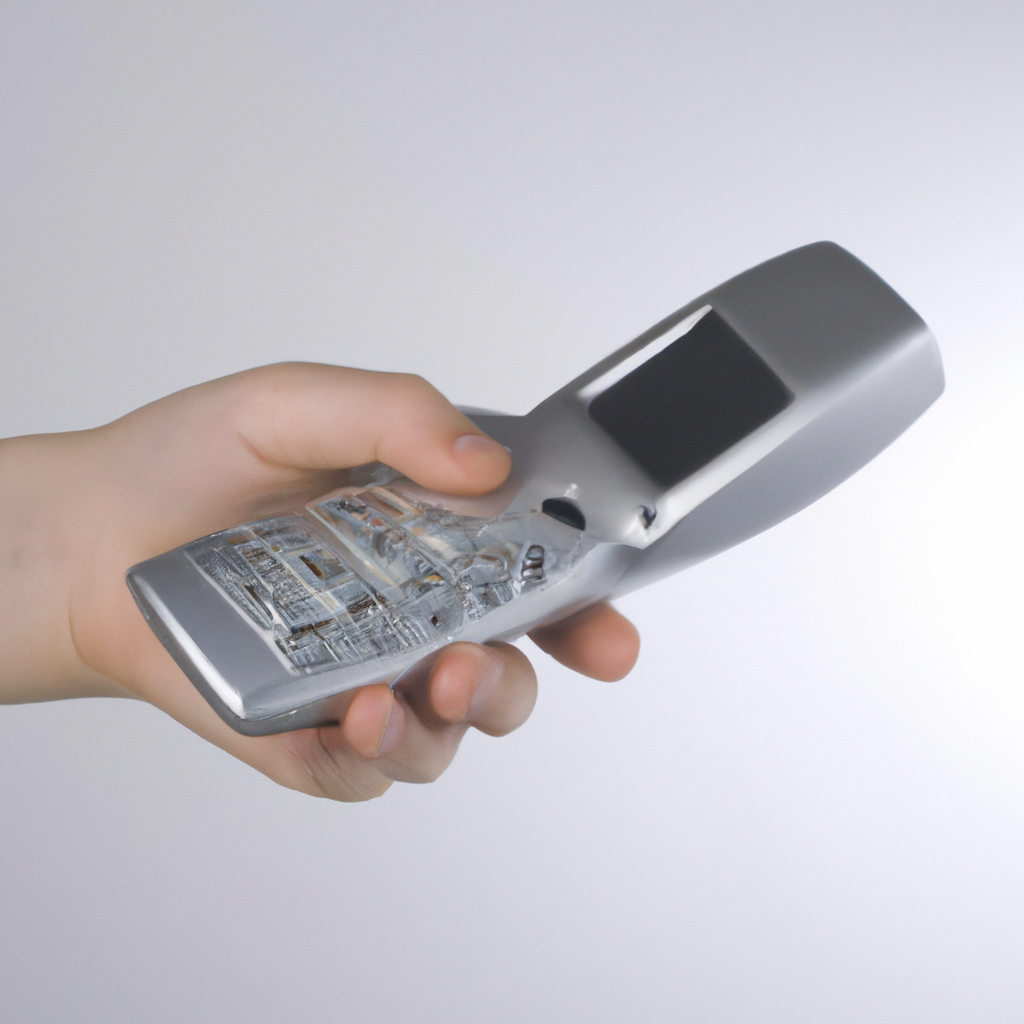}}}%
    \caption{Use a telephone Example}%
    \label{fig:example4}%
\end{figure}

\begin{figure}%
    \centering
    \subfloat[\centering TSI]{{\includegraphics[width=.4\linewidth]{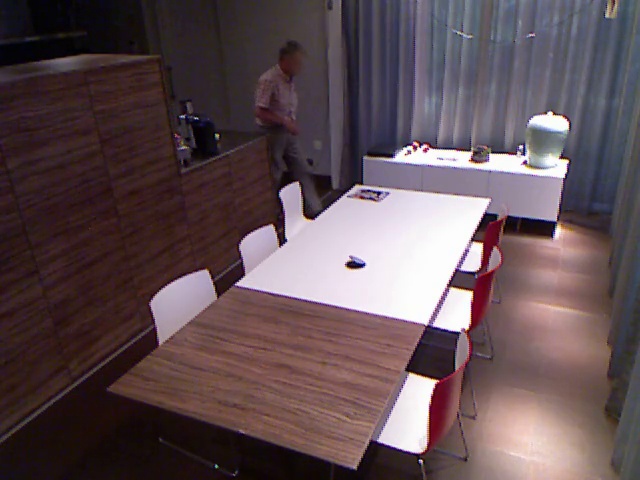}}}%
    \qquad
    \subfloat[\centering DALL·E]{{\includegraphics[width=.4\linewidth]{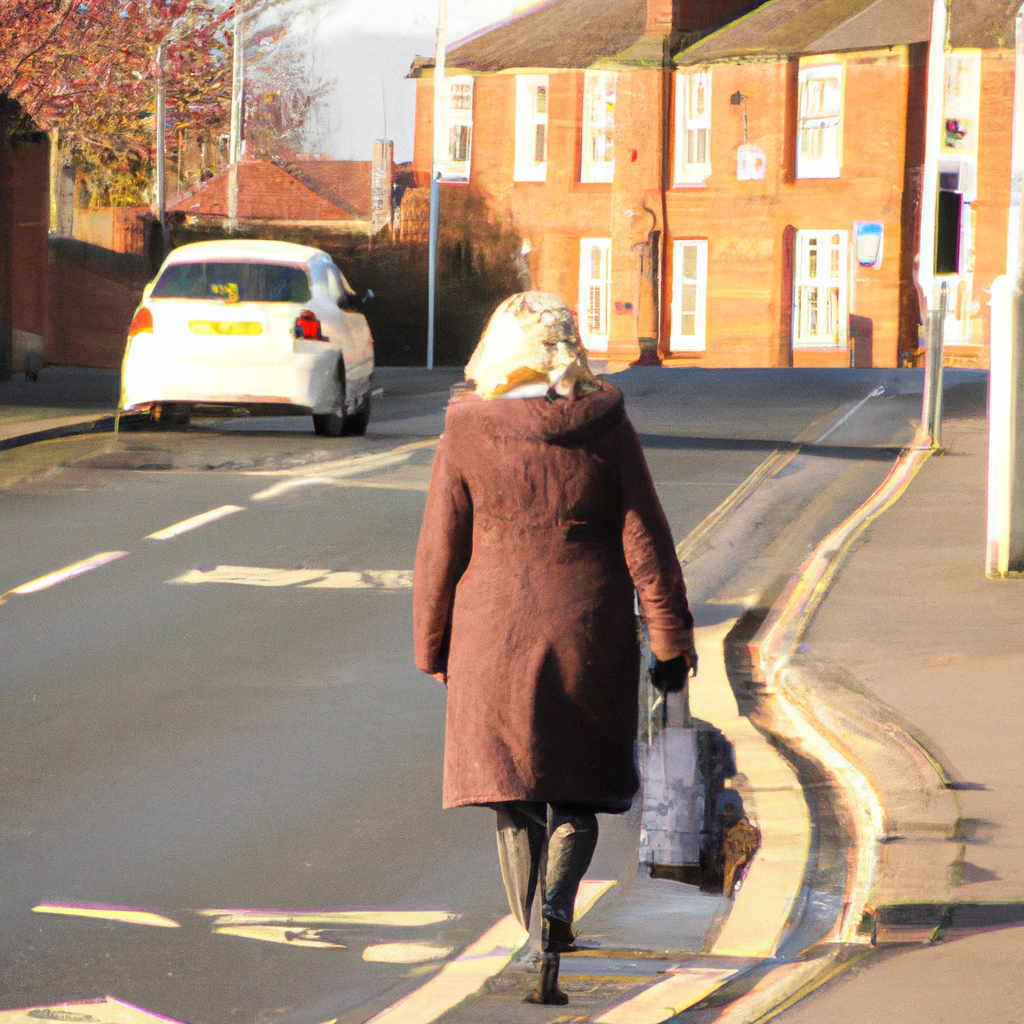}}}%
    \caption{Walking}%
    \label{fig:example5}%
\end{figure}

\begin{figure}%
    \centering
    \subfloat[\centering TSI]{{\includegraphics[width=.4\linewidth]{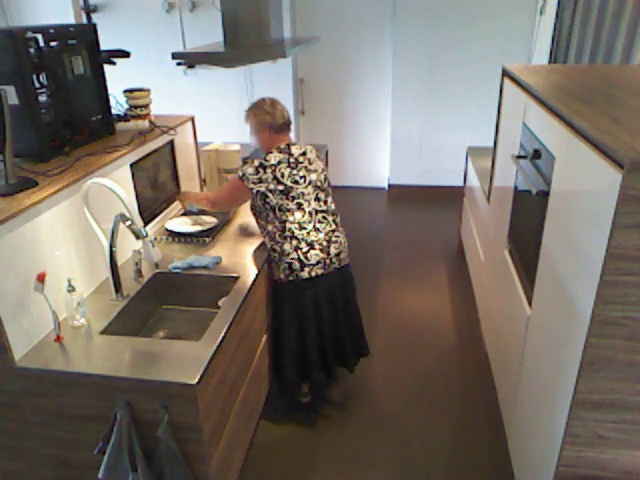}}}%
    \qquad
    \subfloat[\centering DALL·E]{{\includegraphics[width=.4\linewidth]{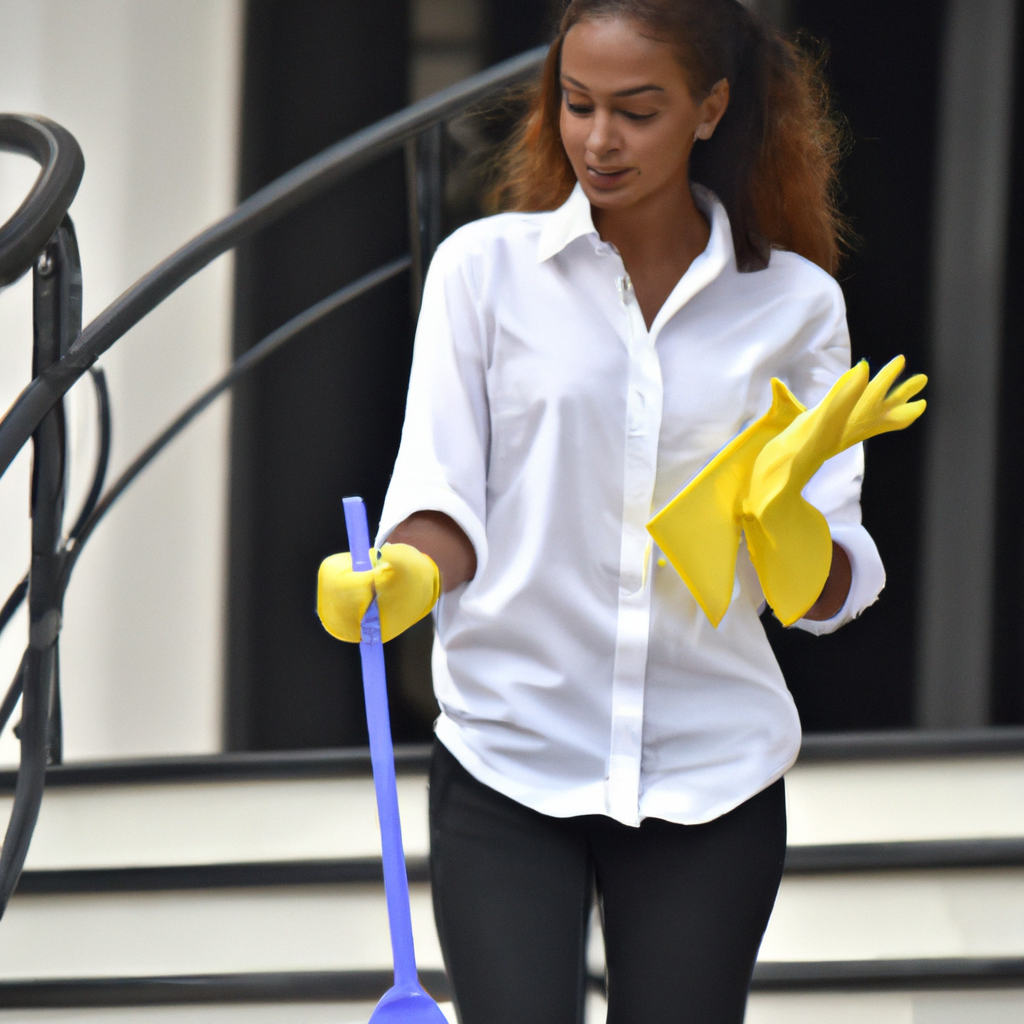}}}%
    \caption{Clean Up Example}%
    \label{fig:example6}%
\end{figure}

\begin{figure}%
    \centering
    \subfloat[\centering TSI]{{\includegraphics[width=.4\linewidth]{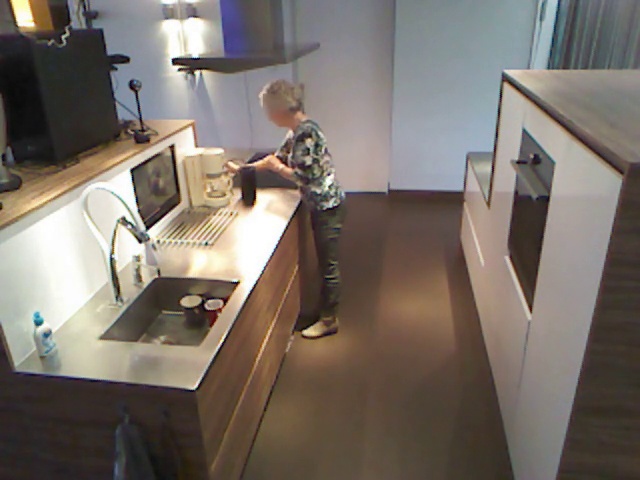}}}%
    \qquad
    \subfloat[\centering DALL·E]{{\includegraphics[width=.4\linewidth]{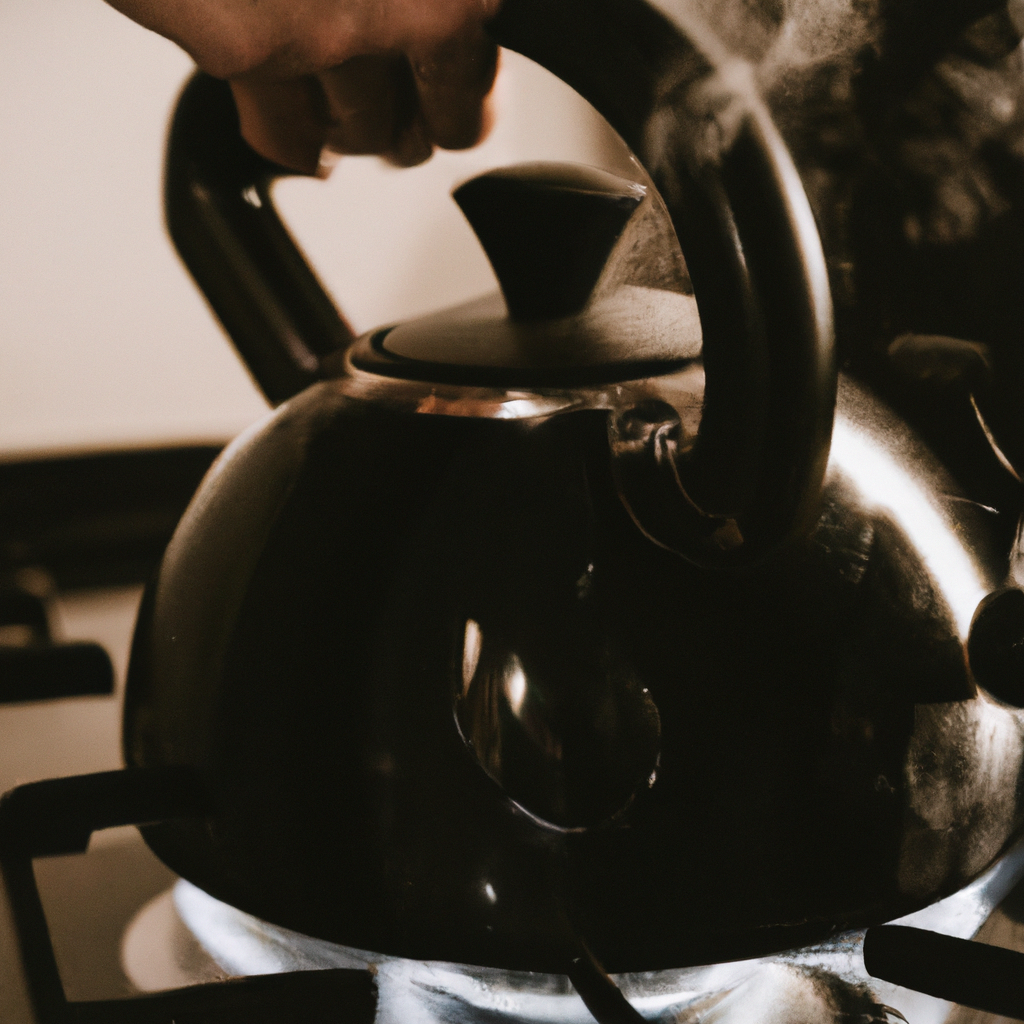}}}%
    \caption{Boiling Water in a Kettle Example}%
    \label{fig:example7}%
\end{figure}

\begin{figure}%
    \centering
    \subfloat[\centering TSI]{{\includegraphics[width=.4\linewidth]{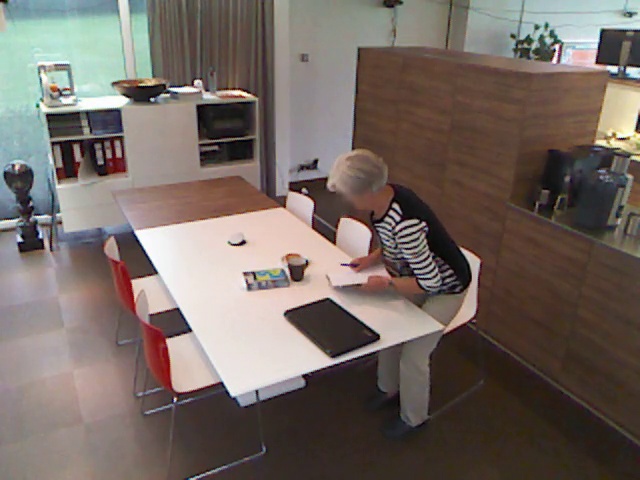}}}%
    \qquad
    \subfloat[\centering DALL·E]{{\includegraphics[width=.4\linewidth]{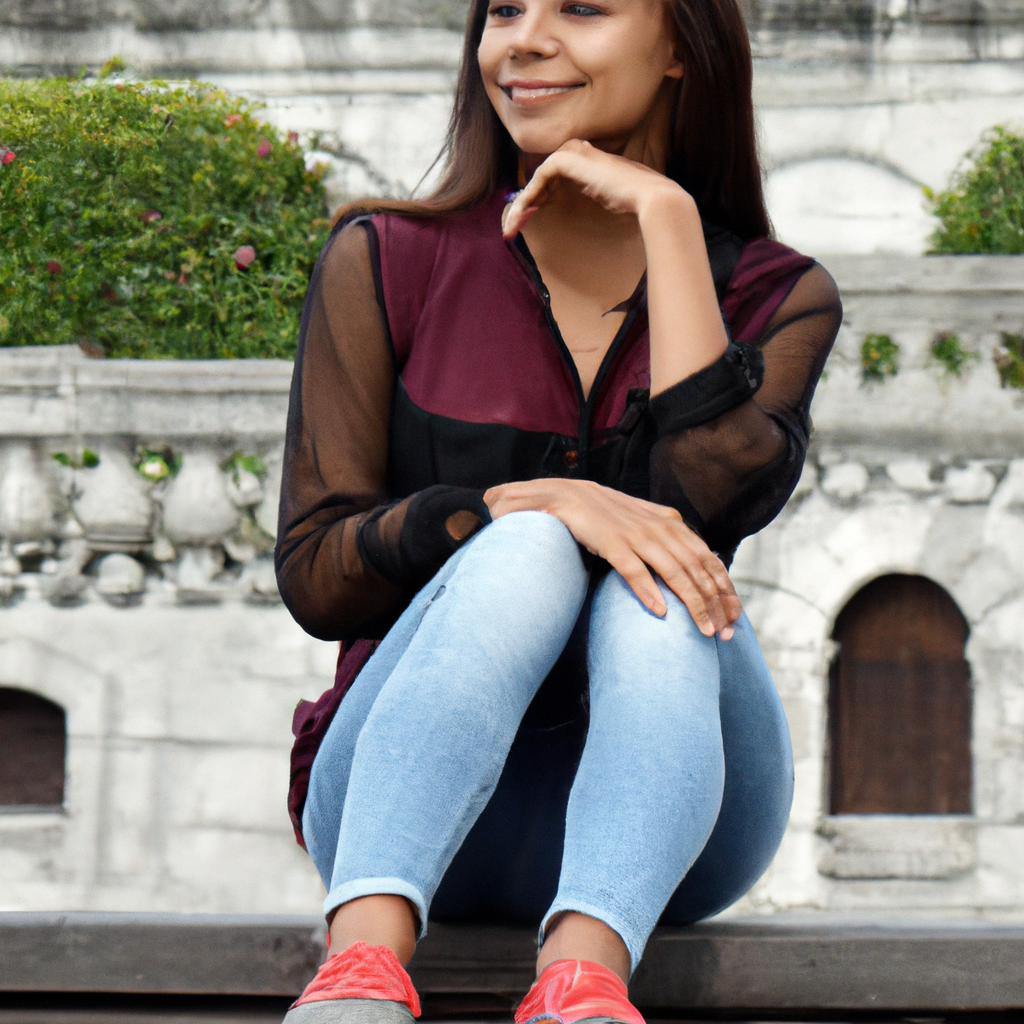}}}%
    \caption{Sit Down Example}%
    \label{fig:example8}%
\end{figure}

\begin{figure}%
    \centering
    \subfloat[\centering TSI]{{\includegraphics[width=.4\linewidth]{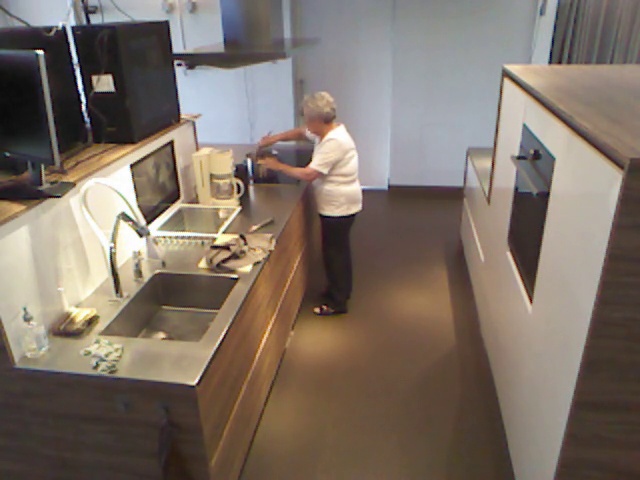}}}%
    \qquad
    \subfloat[\centering DALL·E]{{\includegraphics[width=.4\linewidth]{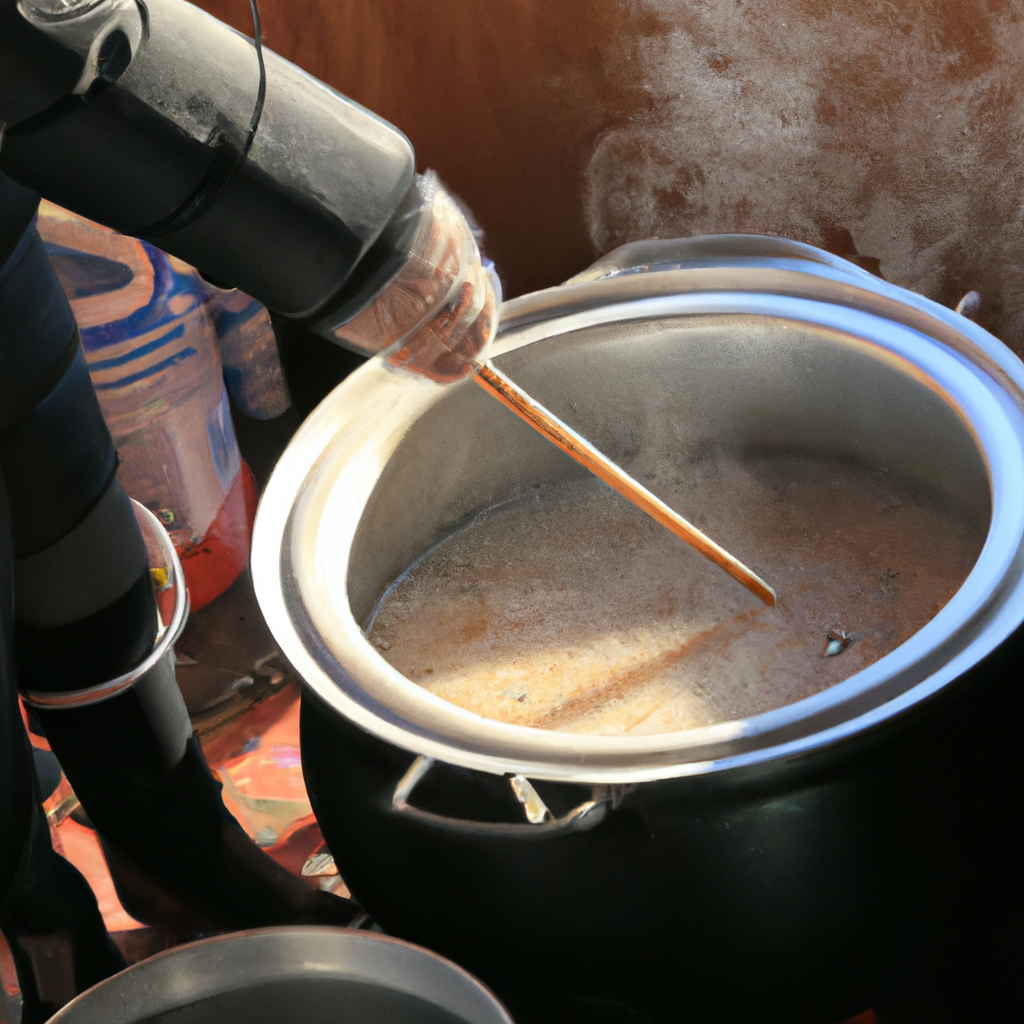}}}%
    \caption{Stirring The Pot Examlpe}%
    \label{fig:example9}%
\end{figure}

We show additional examples of TSI images and DA·E generated images for some actions in Figures~\ref{fig:example1},~\ref{fig:example2},~\ref{fig:example3},~\ref{fig:example4},~\ref{fig:example5},~\ref{fig:example6},~\ref{fig:example7},~\ref{fig:example8},~\ref{fig:example9}.
\end{document}